\documentclass[10pt,twocolumn,letterpaper]{article}

\usepackage[pagenumbers]{cvpr} % To force page numbers, e.g. for an arXiv version

\usepackage[dvipsnames]{xcolor}

\usepackage{graphicx}
\usepackage{amsmath}
\usepackage{amssymb}
\usepackage{subcaption,booktabs}
\usepackage{verbatim}
\usepackage{colortbl}
\usepackage[dvipsnames]{xcolor}
\usepackage{multirow}
\usepackage{setspace}
\usepackage{tabularx}
\usepackage{diagbox}
\usepackage{tikz}
\usepackage{color}

\definecolor{cvprblue}{rgb}{0.21,0.49,0.74}
\usepackage[pagebackref,breaklinks,colorlinks,citecolor=cvprblue]{hyperref}

\def\paperID{8362} % *** Enter the Paper ID here
\def\confName{CVPR}
\def\confYear{2024}

\title{CVT-\textit{x}RF: Contrastive In-Voxel Transformer for 3D Consistent \\ Radiance Fields from Sparse Inputs}

\author{
\begin{tabular}[t]{@{}c@{}}
Yingji Zhong$^1$ \quad Lanqing Hong$^2$ \quad Zhenguo Li$^2$ \quad Dan Xu$^1$
\end{tabular}\\[1ex]
\begin{tabular}[t]{@{}c@{}}
$^1$The Hong Kong University of Science and Technology \quad $^2$Huawei Noah's Ark Lab
\end{tabular}\\[0.5ex]
{\tt\small $\{$yzhongbn,danxu$\}$@cse.ust.hk, $\{$honglanqing,li.zhenguo$\}$@huawei.com}
}

\begin{document}

\twocolumn[{%
\renewcommand\twocolumn[1][]{#1}%
\maketitle
\begin{center}
    \centering
    \vspace{-12pt}
    \includegraphics[width=\textwidth]{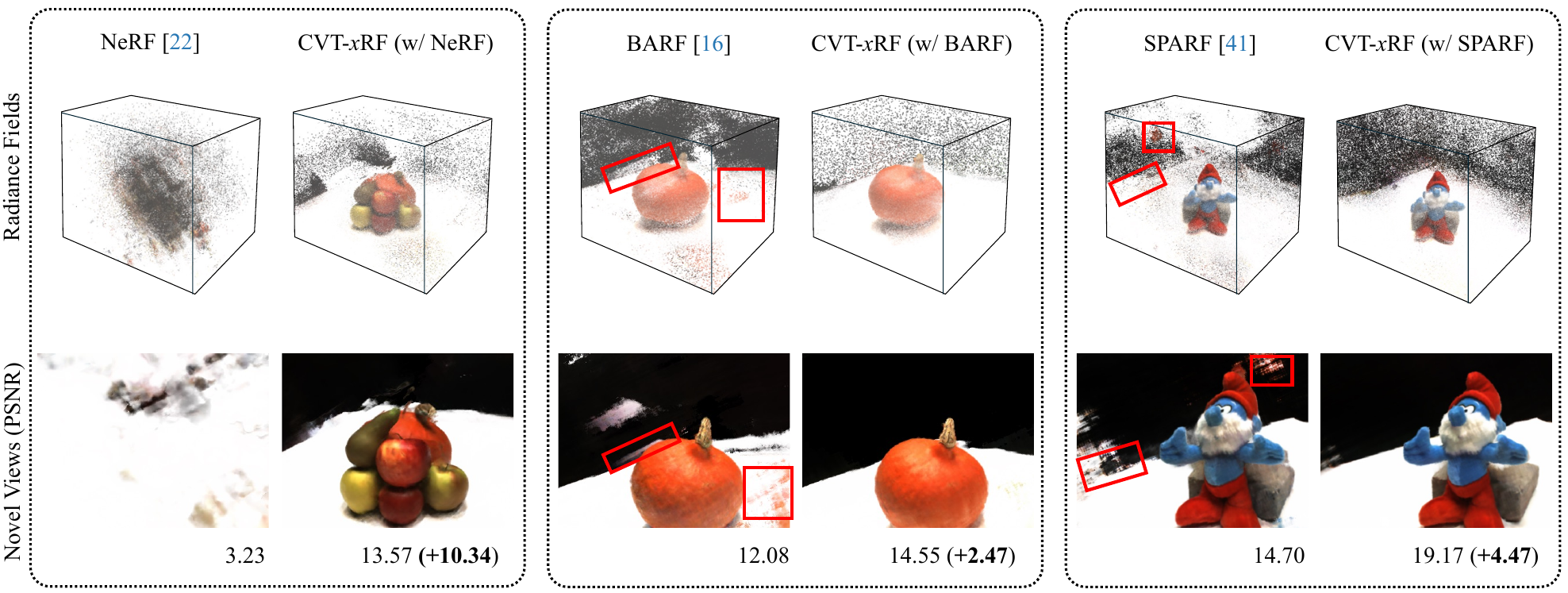}
    \vspace{-20pt}
    \captionof{figure}{Qualitative illustration of learned 3D radiance fields and rendered images of the proposed CVT (Contrastive In-Voxel Transformer)-\textit{x}RF upon three baselines trained from sparse inputs of three views. Our CVT-\textit{x}RF can significantly improve all the baselines. The radiance fields show different levels of 3D inconsistencies (marked in red boxes for BARF and SPARF), which result in failures or artifacts in rendered images. With CVT, we can obtain radiance fields of better 3D consistency and render images of much higher quality. 
    %Moreover, the CVT is only used for training and does not bring any extra overhead in testing. 
    }\label{fig:teaser}
\end{center}%
}]

\begin{abstract}
\vspace{-10pt}
Neural Radiance Fields (NeRF) have shown impressive capabilities for photorealistic novel view synthesis when trained on dense inputs. However, when trained on sparse inputs, NeRF typically encounters issues of incorrect density or color predictions, mainly due to insufficient coverage of the scene causing partial and sparse supervision, thus leading to significant performance degradation. While existing works mainly consider ray-level consistency to construct 2D learning regularization based on rendered color, depth, or semantics on image planes,
in this paper we propose a novel approach that models 3D spatial field consistency to improve NeRF's performance with sparse inputs. Specifically, we first adopt a voxel-based ray sampling strategy to ensure that the sampled rays intersect with a certain voxel in 3D space. We then randomly sample additional points within the voxel and apply a Transformer to infer the properties of other points on each ray, which are then incorporated into the volume rendering. By backpropagating through the rendering loss, we enhance the consistency among neighboring points. Additionally, we propose to use a contrastive loss on the encoder output of the Transformer to further improve consistency within each voxel. Experiments demonstrate that our method yields significant improvement over different radiance fields in the sparse inputs setting, and achieves comparable performance with current works. The project page for this paper is available at \href{https://zhongyingji.github.io/CVT-xRF/}{https://zhongyingji.github.io/CVT-xRF}.  
\end{abstract}    
\vspace{-15pt}
\section{Introduction}
\noindent Representing and modeling 3D properties of scenes is crucial for a wide range of real-world applications, such as autonomous driving, robotic navigation, and 3D content generation. In recent years, 
implicit neural scene representations~\cite{park2019deepsdf,mescheder2019occupancy,sitzmann2019scene,niemeyer2020differentiable,mildenhall2020nerf} have shown impressive abilities to model 3D geometry and appearance in a continuous manner. Among these approaches, Neural Radiance Fields (NeRF)~\cite{mildenhall2020nerf} have emerged as a powerful representation for complex scenes. When the NeRF model is optimized with multi-view inputs, high-fidelity images can be synthesized from unseen novel views~\cite{barron2021mip,verbin2022ref,liu2020neural,barron2022mip,muller2022instant}. 

Despite the significant progress achieved, NeRF has a notable limitation in that it typically requires dense inputs for training its Multi-Layer Perception (MLP). While if only sparse training inputs are provided, because of missing view supervision, NeRF tends to learn a degenerate scene representation that 
fails to accurately model the physical properties (\emph{i.e.}, radiance distributions) of the entire scene, thus resulting in large radiance ambiguities~\cite{zhang2020nerf++}, as can be observed from Fig.~\ref{fig:teaser}. To address this issue, several works have attempted to regularize NeRF during training with different constraints or priors, including sparsity~\cite{kim2022infonerf} and 2D spatial consistency~\cite{niemeyer2022regnerf}, additional depth supervision~\cite{deng2022depth,prinzler2023diner,roessle2022dense}, and semantic alignment~\cite{jain2021putting} or matching~\cite{truong2023sparf} utilizing off-the-shelf pre-trained models.
These existing works have achieved important improvements to this problem. However, they primarily focus on ray-level consistency based on the rendered color and depth, or semantics on 2D image planes, while 3D spatial field consistency is not explicitly modeled. The 3D spatial field consistency reflects a natural phenomenon that the radiance field is spatially consistent, \emph{i.e.}, 3D points physically close or semantically related tend to exhibit similar radiance properties. 
In these existing works, this crucial 3D field consistency can only be \emph{indirectly} regularized through the gradients from the 2D-level regularization onto sampled ray points, making it challenging to effectively model the correlation of radiances among 3D points. 
As also shown in Fig.~\ref{fig:teaser}, the learned radiance fields from sparse inputs of three baselines, \emph{i.e.}, NeRF~\cite{mildenhall2020nerf}, BARF~\cite{lin2021barf}, and SPARF~\cite{truong2023sparf}, exhibit different levels of inconsistency in 3D space, resulting in failures or artifacts in rendered 2D images. 

To explicitly model and learn the aforementioned crucial 3D spatial field consistency, in this paper, we propose a Contrastive In-Voxel Transformer (CVT) structure to implement the 3D field consistency in the sparse inputs setting. 
As illustrated in Fig.~\ref{fig:teaser}, CVT can be flexibly integrated into various baselines, largely boosting the consistency in both 3D radiance fields and 2D rendered images. We denote our method as CVT-\textit{x}RF, where \textit{x} indicates that our CVT structure can be plugged into different baseline radiance fields for sparse-view scene modeling.~Our proposed CVT-\textit{x}RF comprises three main components that work seamlessly to achieve this goal.
More specifically, (i) the first component is a voxel-based ray sampling strategy. In detail, during training, we first select multiple voxels in 3D space. For each selected voxel, we sample rays that intersect with it, which ensures that the 3D points on the rays within the voxel share similar radiance properties.~(ii) The second component of CVT-\textit{x}RF is a local implicit constraint that is based on an In-Voxel Transformer~\cite{vaswani2017attention}. Specifically, for each ray, two distinct sets of 3D points are sampled within the same voxel: one set of points is randomly sampled in the 3D voxel, while the other set of points is sampled along the ray. Since both sets of points are within the same voxel, their radiance properties can be closely correlated. We thus leverage the Transformer to implicitly model the correlation of 3D-point radiances. The Transformer's encoder and decoder take the two sets of points as inputs, respectively. The encoder learns representations of neighboring 3D points; the decoder learns the correlation between neighboring points and ray points, and outputs radiances of the ray points for volume rendering of the ray.~(iii) The third component of CVT-\textit{x}RF is a global explicit constraint in the form of a voxel contrastive regularization. During training, multiple voxels in the 3D scene are sampled, and the contrastive regularization is designed to learn field consistency among positive 3D points (within voxels) and negative 3D points (across voxels). CVT-\textit{x}RF brings significant improvements over different baselines and achieves state-of-the-art performances on multiple challenging benchmarks. 
In summary, our main contributions are as follows: 
\begin{itemize}
    \item We introduce a novel 3D spatial field consistency mechanism for effectively regularizing the learning of radiance fields from sparse inputs.
    \item We propose a Contrastive In-Voxel Transformer (CVT) structure to implement 3D field consistency learning, which is constructed with three key components, \textit{i.e.}, voxel-based ray sampling, local implicit constraint, and global explicit constraint. The CVT structure can be flexibly applied to different baselines.
    \item Our experiments extensively demonstrate that our method brings significant gains over different strong baselines, \textit{e.g.}, on DTU 3-view, our CVT-\textit{x}RF brings 7.45, 0.95, 1.20 PSNR improvements upon NeRF~\cite{mildenhall2020nerf},  BARF~\cite{lin2021barf} and SPARF~\cite{truong2023sparf}, respectively.
\end{itemize}

\begin{figure*}[t]	\includegraphics[width=1.0\linewidth]{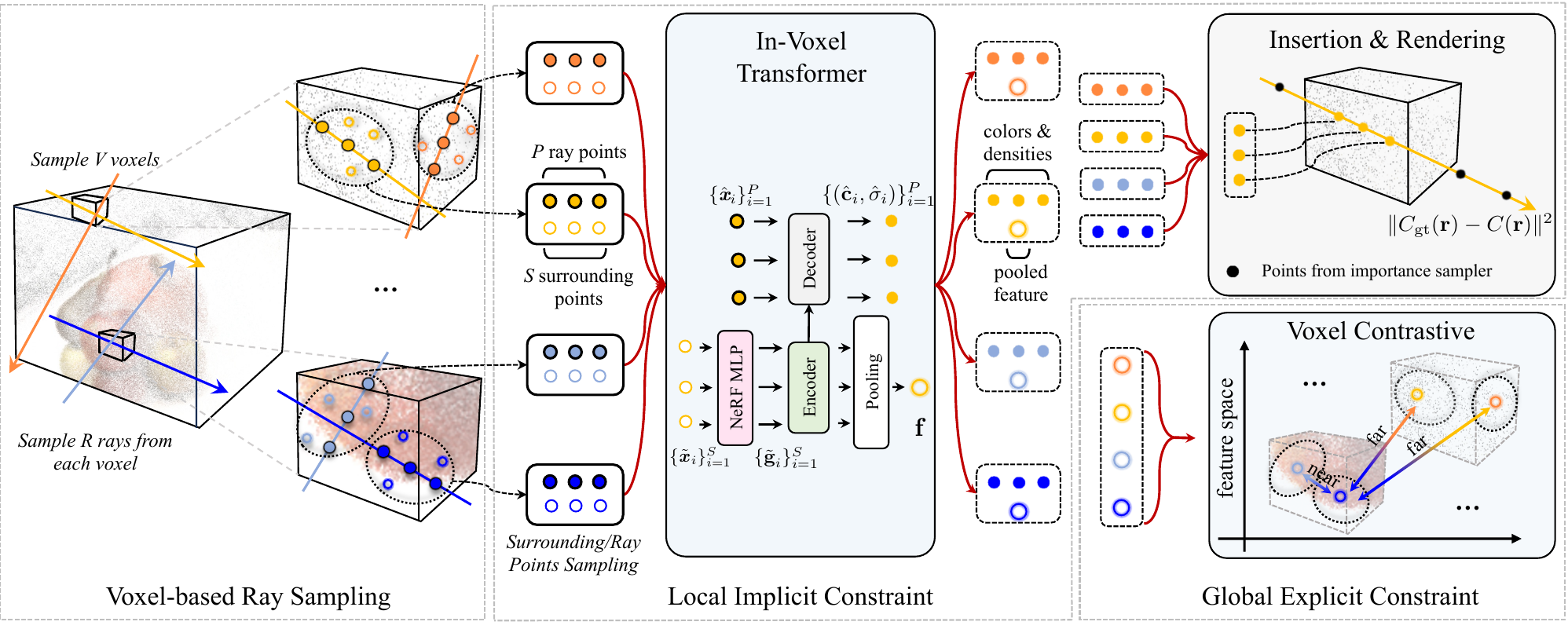}
    \vspace{-18pt}
	\caption{Illustration of the proposed CVT (Contrastive In-Voxel Transformer)-\textit{x}RF for learning radiance fields from sparse inputs. It consists of three parts, \emph{i.e.}, a voxel-based ray sampling strategy, a local implicit constraint module, and a global explicit constraint module. For simplicity, two voxels are shown, along with two rays for each. The local implicit constraint is implemented by a light-weight In-Voxel Transformer which infers colors and densities of ray points by interacting with surrounding 3D points. The ray points are then inserted among the points from the importance sampler for rendering. The global explicit constraint is conducted by a voxel contrastive regularization, which regularizes the radiance properties between points in a voxel to be more similar than that of points across voxels. 
 }
\label{fig:method}
\vspace{-13pt}
\end{figure*}
\section{Related Work}
\noindent \textbf{Neural scene representation.} 
Compared to discretized representations~\cite{sitzmann2019deepvoxels,wu20153d,tatarchenko2017octree,groueix2018papier,qi2017pointnet,qi2017pointnetpp}, neural scene representation~\cite{park2019deepsdf,mescheder2019occupancy} excels in modeling the continuous shape and appearance. With differentiable rendering~\cite{sitzmann2019scene,niemeyer2020differentiable,mildenhall2020nerf}, the model can be trained on posed images. Among them, Neural Radiance Fields (NeRF)~\cite{mildenhall2020nerf} have gained increasing attention in recent years. It achieves impressive results on novel view synthesis with complex scenes~\cite{barron2021mip,verbin2022ref,liu2020neural,muller2022instant}. 
Besides, NeRF has also shown impressive results on other applications~\cite{kosiorek2021nerf,niemeyer2021giraffe,schwarz2020graf,yu2022monosdf,wang2021neus,long2022sparseneus}. 
\par\noindent \textbf{Novel view synthesis from sparse inputs.} One major drawback of NeRF is that it might learn degenerate representations when given sparse inputs~\cite{zhang2020nerf++,niemeyer2022regnerf}.
To address this problem, two lines of research have emerged. 

{The first line of research} aims to learn a generalizable radiance field by pre-training the MLP on multi-view datasets and then fine-tuning it with sparse inputs from a target scene. For example, PixelNeRF~\cite{yu2021pixelnerf} and IBRNet~\cite{wang2021ibrnet} augment the input of MLP with features projected from a CNN feature map. MVSNeRF~\cite{chen2021mvsnerf} builds a 3D volume by warping CNN features and augments the input of the MLP with features from this volume. Although these methods show promising results, they often require multi-view datasets for pre-training, which are not always available, and their performance may drop when given sufficient inputs due to domain differences. 
There are also works focusing on training NeRF with a single image~\cite{watson2022novel,zhou2023sparsefusion,gu2023nerfdiff}. Their methods mainly rely on generative models~\cite{ho2020denoising,rombach2022high} to synthesize images of novel views. Our setting differs from theirs in that the given images can cover the scene from multiple viewpoints.

{The second line of research} utilizes regularization techniques during training. DS-NeRF~\cite{deng2022depth} aligns the density distribution of each ray with the depth supervision which is available from structure-from-motion. A similar method is also applied in DDP~\cite{roessle2022dense}, DINER~\cite{prinzler2023diner} and SparseNeRF~\cite{wang2023sparsenerf}. DietNeRF~\cite{jain2021putting} regularizes the semantic consistency among images from arbitrary views in the embedding space of CLIP~\cite{radford2021learning}. 
InfoNeRF~\cite{kim2022infonerf} applies sparsity regularization by minimizing the entropy of each ray density. RegNeRF~\cite{niemeyer2022regnerf} leverages a 2D consistency loss on depths and colors of image patches to impose that neighboring pixels have similar geometry and appearance. However, in this paper, we explore modeling 3D local and global spatial consistency for optimizing NeRFs from sparse inputs. 
Compared with the 2D consistency utilized in RegNeRF~\cite{niemeyer2022regnerf}, 
3D consistency is a stronger regularization which can 
directly regularizes 3D spatial neighboring regions to learn consistent physical radiance properties. The most relevant method with ours is Nerfbusters~\cite{warburg2023nerfbusters}, which refines local regions to ensure consistency by a data-driven diffusion prior. In contrast, our method applies a contrastive in-voxel Transformer structure to implement 3D consistency from both local and global perspectives without using any off-the-shelf pre-trained models and external priors.

\vspace{3pt}

% \vspace{-10pt}
\section{The Proposed CVT-\textit{x}RF}
\noindent In this work, we propose to use 3D spatial field consistency to regularize radiance fields when training from sparse inputs. Because of the sparse supervision, it is critically important to handle the 3D field consistency in radiance field learning, \emph{i.e.}, neighboring regions in 3D space having similar physical properties, \textit{e.g.}, density and color. Our proposed CVT-\textit{x}RF for implementing 3D field consistency learning is illustrated in Fig.~\ref{fig:method}. Our CVT-\textit{x}RF comprises three major components, which are a voxel-based ray sampling strategy (Sec.~\ref{sec:method_sample}), a local implicit constraint module based on a designed light-weight In-Voxel Transformer (Sec.~\ref{sec:method_implicit}), and a global explicit constraint module based on a voxel contrastive regularization (Sec.~\ref{sec:method_explicit}). We elaborate on them after a brief review of NeRF in Sec.~\ref{sec:method_pre} as follows. 

\subsection{Preliminary}
\label{sec:method_pre}
\noindent Neural Radiance Field (NeRF) adopts an  MLP network to represent a scene, which maps the 3D coordinate $\textit{\textbf{x}}=(x, y, z)$ and its 2D view direction $\textbf{d}=(\theta, \phi)$ to a pair of radiance property values, \emph{i.e.}, $(\textbf{c}, \sigma)$, where $\textbf c$ and $\sigma$ represent the color and the density, respectively. To facilitate subsequent discussions, we additionally extract a feature vector $\textbf{g}$ from the layer of the MLP that predicts the density. The aforementioned process can be formulated as: 
\begin{equation}\label{eq:mlp}
\setlength{\abovedisplayskip}{6pt}
\setlength{\belowdisplayskip}{6pt}
	(\textbf c, \sigma, \textbf{g}) = \rm MLP(\gamma(\textit{\textbf x}), \gamma(\textbf d)),  
\end{equation}
where $\gamma(\cdot)$ refers to a positional encoding. The color of each ray is calculated through volumetric rendering, which accumulates colors of $N$ points sampled from a uniform or an importance sampler. This process can be formulated as: 
$C(\textbf r)=\sum_{i=1}^N T_i(1-\rm exp(-\sigma_{i}\delta_{i}))\textbf{c}_i$, 
where $\delta_i$ refers to the distance between two adjacent samples and $T_i$ is the accumulated transmittance calculated by: $T_i=\rm exp(-\sum_{j=1}^{i-1}\sigma_j \delta_j)$. 
During training, random rays are sampled along with their corresponding ground truth colors. Their colors are rendered by volume rendering and the MLP is learned via supervision from a mean squared error as: 
\begin{equation}\label{eq:mse_loss}
\setlength{\abovedisplayskip}{6pt}
\setlength{\belowdisplayskip}{6pt}
\mathcal L_{\rm mse} = \sum_{\textbf{r}}\Vert C_{\rm gt}(\textbf r) - C(\textbf r)\Vert ^ 2, 
\end{equation}
where $C_{\rm gt}(\textbf r)$ denotes the ground truth color.~Though NeRF achieves impressive results given dense inputs, it cannot recover correct geometry and appearance from sparse inputs and fails to synthesize high-quality images of novel views. 

\vspace{-6pt}
\subsection{Voxel-based Ray Sampling Strategy}\label{sec:vbr}
\label{sec:method_sample}
\noindent Due to the sparse-view supervision, 3D spatial field consistency is not guaranteed in the learned NeRF representation. The consistency indicates that neighboring regions in 3D space have similar radiance properties. However, defining appropriate neighboring regions remains a challenge. We propose a reasonable hypothesis that regions within a small voxel in 3D space are likely to present similar properties, 
and our experiments in supplementary materials demonstrate that this hypothesis leads to significant gains across a wide range of voxel sizes.
To implement this hypothesis, we uniformly divide the scene into voxels with an equal size. Since we have access to training images and their corresponding camera parameters, we can record the voxels that each ray intersects with. Each voxel can then store multiple rays that intersect with it. We introduce a voxel-based ray sampling strategy as shown in Fig.~\ref{fig:method}, which supports the local and global field consistency learning proposed in Sec.~\ref{sec:method_implicit} and Sec.~\ref{sec:method_explicit}. Our strategy starts with a sampling of $V$ voxels, denoted as $\{\textbf{V}_i\}_{i=1}^V$. It then samples $R$ rays from each voxel and returns $\{\textbf{R}_i\}_{i=1}^V$, where each $\textbf R_i$ refers to a set of $R$ rays sampled from a voxel $\textbf V_i$. Using this strategy, we can rewrite the supervision of Eq.~\eqref{eq:mse_loss} as: 
\begin{equation}\label{eq:mse_loss_vbr}
\setlength{\abovedisplayskip}{4pt}
\setlength{\belowdisplayskip}{4pt}
\mathcal L_{\rm mse} = \sum_{i=1}^V \sum_{\textbf r \in \textbf R_i} \Vert C_{\rm gt}(\textbf r) - C(\textbf r)\Vert ^ 2, 
\end{equation}
and the batch size of training rays is thus $V\times R$. 

\begin{figure}[t]
	\centering
	\includegraphics[width=0.99\linewidth]{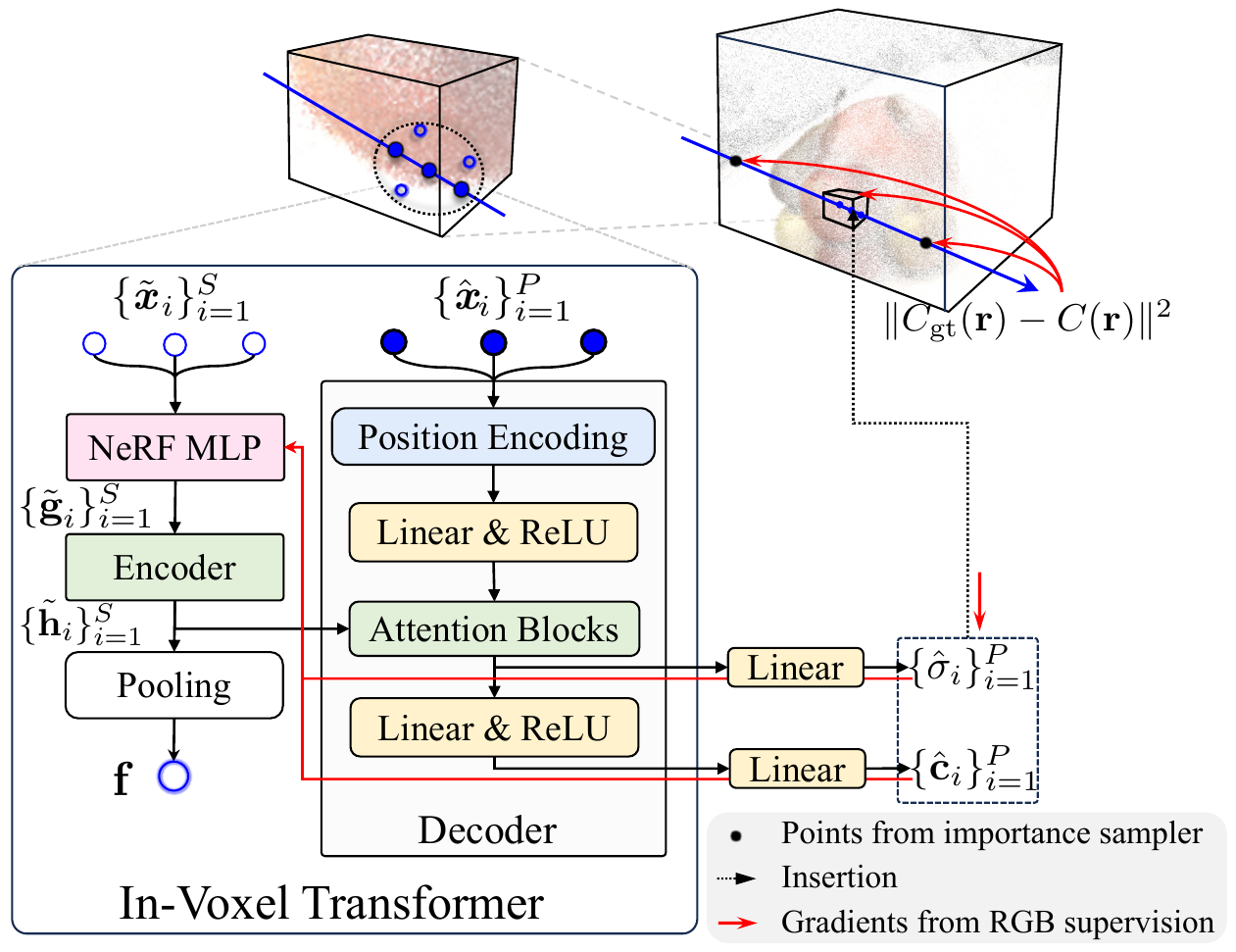}
 \vspace{-6pt}
\caption{The proposed local implicit constraint with a light-weight In-Voxel Transformer for 3D field consistency learning. The colors and densities of ray points $\{\hat{\textit{\textbf{x}}}_i\}_{i=1}^P$ are predicted by surrounding points $\{\tilde{\textbf{\textit x}}_i\}_{i=1}^S$ through the Transformer. The predicted colors and densities are inserted into the ray for rendering.  
} \vspace{-13pt}
\label{fig:method_backprop}
\end{figure}

\subsection{Local Implicit Constraint}
\label{sec:method_implicit}
\noindent Now we start to introduce the proposed local implicit constraint for 3D field consistency learning. Considering a voxel $\textbf{V}$, the physical radiance properties of two small regions in this voxel should be similar, which means we can infer the properties of one region through an interaction with the other. In the following, we represent these two regions in the voxel $\textbf V$ by two point sets, which contain the surrounding points and the ray points, respectively. For conciseness, we drop subscripts and only consider a ray $\textbf r$ sampled from the voxel $\textbf V$. 

\noindent \textbf{Surrounding and ray points sampling.}~As shown in Fig.~\ref{fig:method}, in voxel $\textbf V$, we sample $S$ surrounding points $\{\tilde{\textbf{\textit x}}_i\}_{i=1}^S$ for each ray $\textbf r$. Concretely, we firstly obtain two points that the ray intersects $\textbf V$ with, \emph{i.e.}, $\textit{\textbf x}_{\rm in}$ and  $\textit{\textbf x}_{\rm out}$. We then perform sphere sampling using $\mathcal{F}_{\rm sphere{\_}sample}$ with the center located at the midpoint of the two intersecting points as:  
\begin{equation}\label{eq:sur_sample}
\setlength{\abovedisplayskip}{6pt}
\setlength{\belowdisplayskip}{6pt}
	\{\tilde{\textbf{\textit x}}_i\}_{i=1}^S = \mathcal{F}_{\rm sphere{\_}sample}((\textbf{\textit x}_{\rm in}+\textbf{\textit x}_{\rm out})/2, \rm radius, \textit S), 
\end{equation}
where $\rm radius$ is set to $1/n$ of the voxel size.
According to the 3D field consistency discussed, the radiance properties (\emph{i.e.}, colors and densities) of the surrounding points are highly beneficial for inferring those of the ray points. As shown in Fig.~\ref{fig:method}, we sample $P$ ray points $\{\hat{\textit{\textbf{x}}}_i\}_{i=1}^P$ along the ray $\textbf r$ with $\mathcal{F}_{\rm line\_sample}$. Specifically, we randomly sample points along the line segment connecting $\textit{\textbf x}_{\rm in}$ and  $\textit{\textbf x}_{\rm out}$ as: 
\begin{equation}\label{eq:ray_sample}
\setlength{\abovedisplayskip}{6pt}
\setlength{\belowdisplayskip}{6pt}
	\{\hat{\textit{\textbf{x}}}_i\}_{i=1}^P=\mathcal{F}_{\rm line{\_}sample}(\textbf{\textit x}_{\rm in}, \textbf{\textit x}_{\rm out}, \textit P). 
\end{equation}

\par\noindent \textbf{Prediction by a light-weight In-Voxel Transformer.}~After obtaining the surrounding points and ray points, we introduce a light-weight Transformer structure to perform inference of the radiance properties of the ray points based on the surrounding points. The proposed Transformer structure consists of an encoder and a decoder. The encoder is designed to encode the properties of the region containing the surrounding points, while the decoder is responsible for decoding the radiances of the ray points based on the encoded information from the surrounding points. 
\par Fig.~\ref{fig:method_backprop} illustrates the details of the Transformer structure.
The input to the encoder consists of the point features $\textbf{g}$ obtained from the MLP, as depicted in Eq.~\ref{eq:mlp}. Concretely, We forward the coordinates of the surrounding points into MLP and obtain their corresponding features, \textit{i.e.}, $\{\tilde{{\textbf{g}}}_i\}_{i=1}^S=\rm MLP(\{\tilde{\textbf{\textit x}}_i\}_{i=1}^S, \textbf d)$. The encoding procedure is as follows: 
\begin{equation}\label{eq:encoder}
\setlength{\abovedisplayskip}{6pt}
\setlength{\belowdisplayskip}{6pt}
	\{\tilde{\textbf{h}}_i\}_{i=1}^S = {\rm Encoder }(\{\tilde{\textbf{g}}_i\}_{i=1}^S), 
\end{equation}
where {${\rm Encoder}$ consists of self-attention blocks and outputs updated features $\{\tilde{\textbf{h}}_i\}_{i=1}^S$. }To achieve a compact representation, we aggregate $\{\tilde{\textbf{h}}_i\}_{i=1}^S$ by a pooling operation as: 
\begin{equation}\label{eq:pool}
\setlength{\abovedisplayskip}{6pt}
\setlength{\belowdisplayskip}{6pt}
	\mathbf{f} = {\rm Pool}(\{\tilde{\textbf{h}}_i\}_{i=1}^S), 
\end{equation}
where $\rm Pool$ is implemented with a max pooling in practice. $\mathbf{f}$ is now the feature that represents a specific 3D region containing the surrounding points in the voxel. Then, the decoder aims at inferring colors and densities of the ray points from the encoded representation $\{\tilde{\textbf{h}}_i\}_{i=1}^S$ of surrounding points. 
Different from the common practice of NeRF that predicts the densities and colors of points in isolation, the decoder performs the prediction based on the encoded information from the neighboring points.  
The detailed architecture of the decoder is illustrated in Fig.~\ref{fig:method_backprop}. {For $P$ ray points, the decoder firstly transforms their coordinates by position encoding, which is followed by a non-linearity. The self-attention blocks receive both the non-linear output and the encoder output $\{\tilde{\textbf{h}}_i\}_{i=1}^S$, to decode the radiances for the $P$ ray points. The outputs of the attention blocks are directly utilized to predict the densities of the ray points, \textit{i.e.}, $\{\hat{\sigma}_i\}_{i=1}^P$. After applying a non-linearity to the outputs of the attention blocks, the colors, denoted as $\{\hat{\textbf c}_i\}_{i=1}^P$, are predicted. }
The decoding procedure is formulated as: 
\begin{equation}\label{eq:decoder}
\setlength{\abovedisplayskip}{6pt}
\setlength{\belowdisplayskip}{6pt}
	\{(\hat{\textbf c}_i, \hat{\sigma}_i)\}_{i=1}^P = {\rm Decoder} (\{\hat{\textit{\textbf{x}}}_i\}_{i=1}^P, \{\tilde{\textbf{h}}_i\}_{i=1}^S). 
\end{equation}

\noindent \textbf{Insertion and rendering.} After we have obtained the colors and densities of $P$ ray points, we then insert them into the ray. Along with the original $N$ points that are sampled from the importance sampler, we combine the radiances of the $N+P$ points for volume rendering as illustrated in Fig.~\ref{fig:method_backprop}. During training, the gradients from the color rendering loss (Eq.~\ref{eq:mse_loss_vbr}) are backpropagated to the points along the ray, including the ray points. As shown in Fig.~\ref{fig:method_backprop}, the gradients can flow back to the encoder, thus updating the parameters of the MLP. 
Through the interaction between the surrounding points and the ray points in the neighboring regions, our local implicit constraint can largely enhance the 3D field consistency during training.

\begin{figure}[!t]
\vspace{-5pt}
	\centering	\includegraphics[width=1.0\linewidth]{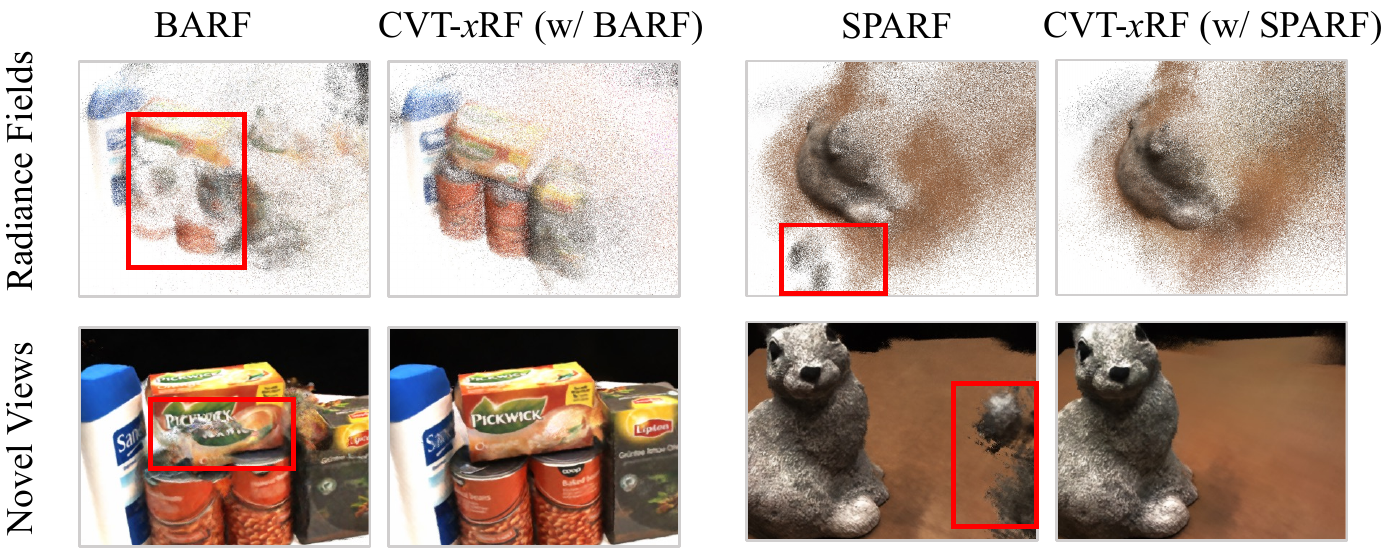}
  \vspace{-16pt}
	\caption{Visualization of learned radiance fields (in 3D) and the corresponding rendering results of two baselines, \emph{i.e.}, BARF~\cite{lin2021barf} and SPARF~\cite{truong2023sparf}, with/without the proposed CVT. 
 } \vspace{-12pt}
	\label{fig:vis_pointcloud}
\end{figure}

\subsection{Global Explicit Constraint}
\label{sec:method_explicit}
\noindent The local implicit constraint enhances the 3D field consistency of neighboring regions by the interaction between surrounding points and ray points.
In this section, we propose a global explicit constraint, which directly enforces the similarity between features of neighboring regions for 3D field consistency. 
After applying the local implicit constraint, we can obtain a pooled feature $\mathbf{f}$ from the features of a set of surrounding points as in Eq.~\eqref{eq:pool}. Thus, each $\mathbf{f}$ can represent % the properties of 
a specific region depicted by the set of surrounding points within a voxel, we denote $\mathbf{f}$ as a region feature. Following the voxel-based ray sampling strategy in Sec.~\ref{sec:method_sample}, we sample $V$ voxels, with $R$ rays from each voxel during training. We can then obtain a set of features from the encoder output of the Transformer, \textit{i.e.}, 
$\{\mathbf{f}_i^{(j)}\}_{i=1:V, j=1:R}$. 

\par
\begin{figure}[t]
	\centering	\includegraphics[width=0.99\linewidth]{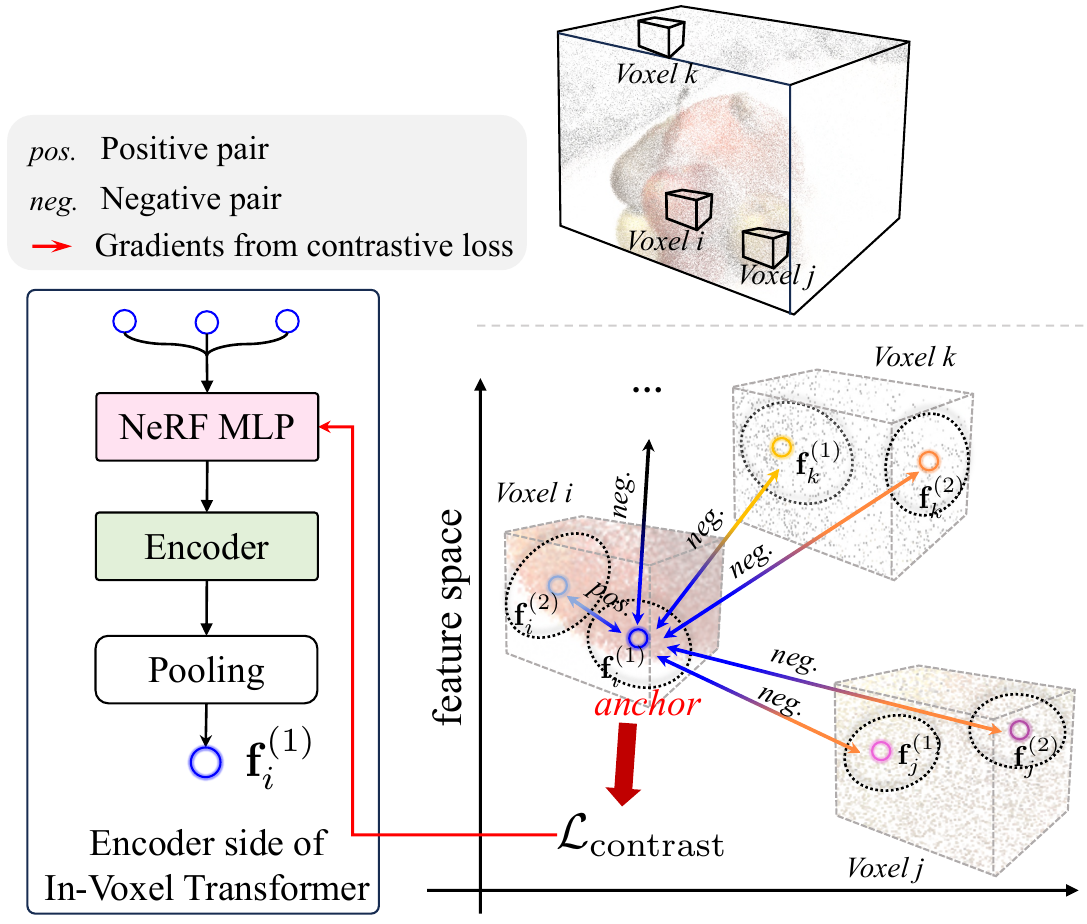}
 \vspace{-6pt}
	\caption{The proposed global explicit constraint for field consistency learning is implemented by a contrastive loss. For each anchor, we select its positive pair in the same voxel, while its negative pairs in other voxels.
    }\vspace{-12pt}
	\label{fig:method_contrast}
\end{figure}

 Inspired by contrastive learning schemes~\cite{sohn2016improved,chen2020simple} that learn discriminative features via optimizing the distance between feature pairs, we propose to use a voxel contrastive loss to further enhance the 3D field consistency. Specifically, for each region feature (an anchor), its distance to other neighboring features from the same voxel (positive pairs) should be smaller than the distance to region features from other voxels (negative pairs). Following Chen~\textit{et al.}~\cite{chen2020simple}, for each anchor region feature, we only select a positive region feature from the same voxel to construct a positive pair, and select all negative region features in other voxels to build its negative pairs. The positive/negative pair selection is illustrated in Fig.~\ref{fig:method_contrast}. The contrastive loss $\mathcal{L}_{\rm contrast}$ regarding $V \times R$ region features can be formulated as: 
\begin{equation}\label{eq:contrastive}
\setlength{\abovedisplayskip}{6pt}
\setlength{\belowdisplayskip}{6pt}
    \begin{aligned}
    &\mathcal{L}_{\rm contrast} = -\sum_{i=1}^V \sum_{j=1}^R \bigg(\frac{\langle \mathbf{f}_i^{(j)}, \mathbf{f}_i^{(pos)} \rangle}{\tau} - \\
    &{\rm log}  \Big ({\rm exp}\big(\frac{\langle \mathbf{f}_i^{(j)}, \mathbf{f}_i^{(pos)} \rangle}{\tau}\big)+\sum_{n=1 \atop n \neq i}^{V}\sum_{k=1}^R {\rm exp}\big(\frac{\langle \mathbf{f}_i^{(j)}, \mathbf{f}_n^{(k)}\rangle}{\tau}\big)\Big)\bigg), 
    \end{aligned}
    \nonumber
\end{equation}
where $\langle \cdot \rangle$ and $\tau$ denote the cosine similarity and the temperature. $\mathbf{f}^{({pos})}$ denotes a positive pair of the anchor, which is randomly selected from the same voxel with the anchor. 
During training, the gradients from the contrastive loss can flow back to the MLP (see Fig.~\ref{fig:method_contrast}) to update its parameters, which makes the MLP network learn more consistent region features, 
% \textit{i.e.}, for the feature of a point, the feature similarity between it and the other point in the same voxel, is higher than that in another voxel. 
leading to better 3D field consistency. 

\vspace{-4pt}
\subsection{Overall Objective}\label{sec:method_obj}
Based on the voxel-based ray sampling strategy, as well as the local implicit and the global explicit constraints, the overall training loss of our CVT-\textit{x}RF can be formulated as: 
\begin{equation}
\setlength{\abovedisplayskip}{6pt}
\setlength{\belowdisplayskip}{6pt}
	\mathcal L = \mathcal L_{\rm mse} + \lambda \mathcal L_{\rm contrast}, 
\end{equation}
where $\lambda$ is a balancing parameter. NeRF commonly uses a coarse-level and a fine-level MLP, which apply a uniform sampling and an importance sampling, respectively. Our proposed CVT-\textit{x}RF is only applied on the fine-level MLP.

\section{Experiments}
\subsection{Datasets and Evaluation}
\noindent \textbf{Datasets.} We evaluate our proposed method on multi-view DTU dataset~\cite{jensen2014large} and Synthetic dataset~\cite{mildenhall2020nerf}.  We report results of 3, 6, and 9 input views for the DTU dataset, while 3 and 8 input views for the Synthetic dataset. 
For more details about the scenes and view selection, we refer readers to the supplementary material.

\par\noindent \textbf{Evaluation.} For quantitative comparison of synthesis results, we report the mean of peak signal-to-noise ratio (PSNR) and structural similarity index (SSIM)~\cite{wang2004image} over different scenes. We refer readers to supplementary material for LPIPS perceptual metric~\cite{zhang2018unreasonable}, geometric mean~\cite{niemeyer2022regnerf}, and the consideration of evaluation on DTU dataset.

\begin{figure}[!t]
	\centering	\includegraphics[width=1.0\linewidth]{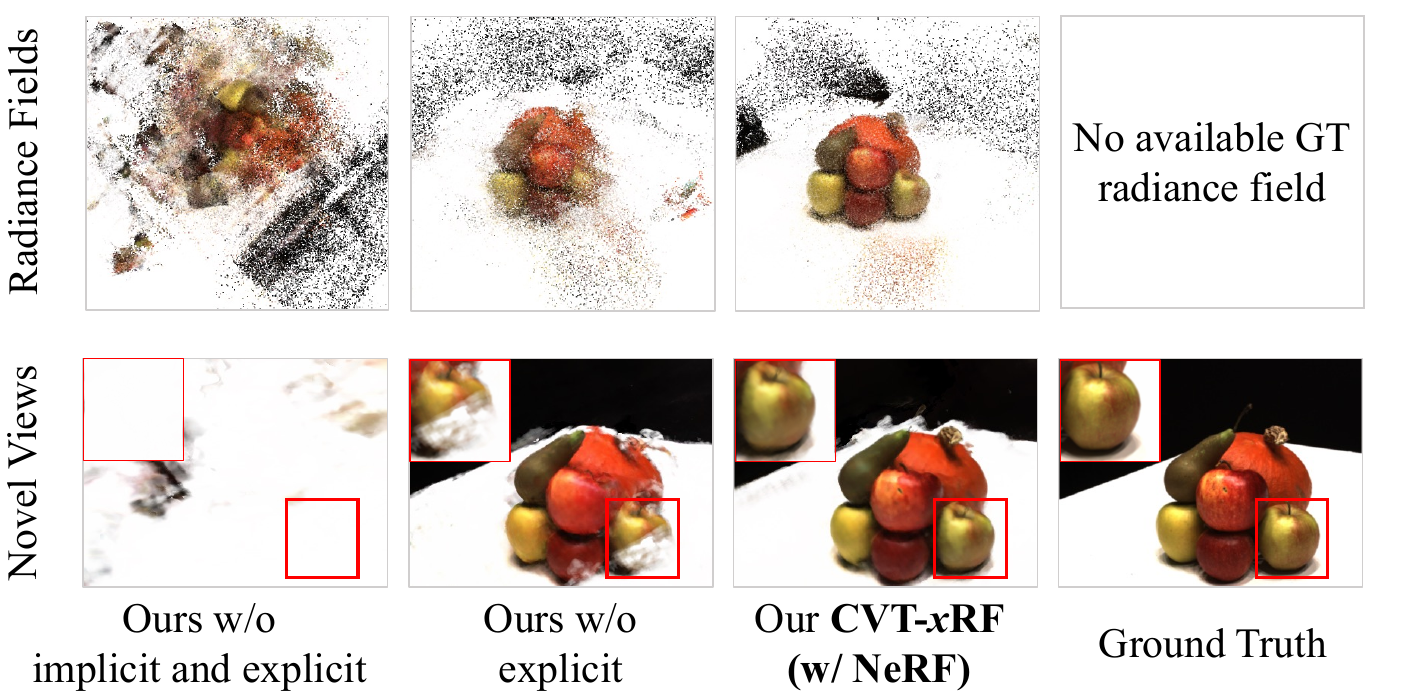}
 \vspace{-15pt}
	\caption{Efficacy of local implicit and global explicit constraints. 
 }
 \vspace{-2pt}
	\label{fig:vis_components}
\end{figure}

\begin{table}[t]
\begin{center}
\resizebox{0.9\linewidth}{!}{
\begin{tabular}{l|cc|cc}
\toprule[1.pt]
\multirow{2}{*}{Methods}                    & \multicolumn{2}{c|}{DTU 3-view}        & \multicolumn{2}{c}{Synthetic 3-view}   \\ %\cline{2-5} 
                                     & PSNR           & SSIM           & PSNR           & SSIM           \\ \midrule[0.5pt]
 NeRF~\cite{mildenhall2020nerf}              & 6.68           & 0.249          & 11.41          & 0.724               \\
CVT-\textit{x}RF (w/ NeRF) & \textbf{14.13} & \textbf{0.518} & \textbf{19.28} & \textbf{0.815} \\
    \quad w/o {implicit and explicit}          & 7.23           & 0.264          & 15.66          & 0.795          \\
  \quad w/o {explicit}       & 11.85          & 0.458          & 18.26          & 0.806          \\
 \bottomrule[1.pt]
\end{tabular}}
\end{center}\vspace{-15pt}\caption{Effectiveness of different parts of the proposed CVT-\textit{x}RF over vanilla NeRF. Implicit and explicit refer to the local implicit constraint and global explicit constraint, respectively. }\vspace{-12pt}\label{table:ablation}
\end{table}

\subsection{Ablation Studies}\vspace{-1pt}
Our CVT-\textit{x}RF consists of three parts, which are voxel-based ray sampling, local implicit constraint, and global explicit constraint. We analyze the effect of these components on the DTU and the Synthetic dataset of 3 input views. 

\begin{table*}[th]
\centering
\resizebox{1\linewidth}{!}{
\begin{tabular}{lc|c|cccccccccccc}
\toprule[1.pt]
\multicolumn{1}{l|}{\multirow{2}{*}{Methods}} & \multicolumn{1}{c|}{\multirow{2}{*}{\textit{Pre.}}} & \multicolumn{1}{c|}{\multirow{2}{*}{Setting}} & \multicolumn{3}{c|}{Full-image PSNR}                                     & \multicolumn{3}{c|}{Full-image SSIM}                                     & \multicolumn{3}{c|}{Object PSNR}                                         & \multicolumn{3}{c}{Object SSIM}                                    \\ %\cline{4-15} 
\multicolumn{1}{l|}{}  &    &           & 3-view                    & 6-view                    & \multicolumn{1}{c|}{9-view}     & 3-view                    & 6-view                    & \multicolumn{1}{c|}{9-view}     & 3-view                    & 6-view                    & \multicolumn{1}{c|}{9-view}     & 3-view                    & 6-view                    & 9-view                    \\ \midrule[0.5pt]
\multicolumn{1}{l|}{PixelNeRF~\cite{yu2021pixelnerf}}  & \multirow{4}{*}{$\times$}       &\multirow{1}{*}{Trained on}   & \textbf{\textcolor{BrickRed}{18.74}}                & 21.02                & \multicolumn{1}{c|}{22.23} & 0.618                & 0.684                & \multicolumn{1}{c|}{0.714} & 16.82                & 19.11                & \multicolumn{1}{c|}{20.40} & 0.695                & 0.745                & 0.768                \\
\multicolumn{1}{l|}{MVSNeRF~\cite{chen2021mvsnerf}}        & &\multirow{1}{*}{DTU and}  & 16.33                & 18.26                & \multicolumn{1}{c|}{20.32} & 0.602                & 0.602                & \multicolumn{1}{c|}{0.735} & 18.63                & 20.70                & \multicolumn{1}{c|}{22.40} & \textcolor{Apricot}{0.769}                & 0.823                & 0.853                \\
\multicolumn{1}{l|}{PixelNeRF ft~\cite{yu2021pixelnerf}}   &   &\multirow{1}{*}{Finetuned}  & 17.38                & \textcolor{Apricot}{21.52}                & \multicolumn{1}{c|}{21.67} & 0.548                & 0.670                & \multicolumn{1}{c|}{0.680} & \textcolor{Apricot}{18.95}                & 20.56                & \multicolumn{1}{c|}{21.82} & 0.710                & 0.753                & 0.781                \\
\multicolumn{1}{l|}{MVSNeRF ft~\cite{chen2021mvsnerf}}    &  &\multirow{1}{*}{per Scene}   & 16.26                & 18.22                & \multicolumn{1}{c|}{20.32} & 0.601                & 0.601                & \multicolumn{1}{c|}{0.736} & 18.54                & 20.49                & \multicolumn{1}{c|}{22.22} & \textcolor{Apricot}{0.769}                & 0.822                & 0.853                \\ \midrule[0.5pt]
\multicolumn{1}{l|}{NeRF~\cite{mildenhall2020nerf}}     & \multirow{5}{*}{$\times$}   &       & 6.68                     & 15.32                     & \multicolumn{1}{c|}{16.29}      & 0.249                      & 0.626                     & \multicolumn{1}{c|}{0.693}      & 7.79                 & 18.23                & \multicolumn{1}{c|}{18.80} & 0.595                & 0.758                & 0.801                \\
\multicolumn{1}{l|}{mip-NeRF~\cite{barron2021mip}}      & &     & 7.64                 & 14.33                & \multicolumn{1}{c|}{20.71} & 0.227                & 0.568                & \multicolumn{1}{c|}{0.799} & 8.68                 & 16.54                & \multicolumn{1}{c|}{23.58} & 0.571                & 0.741                & 0.879                \\
\multicolumn{1}{l|}{RegNeRF~\cite{niemeyer2022regnerf}}  & &\multirow{1}{*}{Optimized}         & 15.33                & 19.10                & \multicolumn{1}{c|}{\textcolor{Apricot}{22.30}} & \textcolor{Apricot}{0.621}                & \textcolor{Apricot}{0.757}                & \multicolumn{1}{c|}{\textcolor{Apricot}{0.823}} & 18.89                & 22.20                & \multicolumn{1}{c|}{24.93} & 0.745                & \textcolor{RoyalBlue}{0.854}                & \textcolor{Apricot}{0.884}                \\

\multicolumn{1}{l|}{FlipNeRF~\cite{seo2023flipnerf}}  & &\multirow{1}{*}{per Scene}     & -                 & -               & \multicolumn{1}{c|}{-} & -                & -                & \multicolumn{1}{c|}{-} & 19.55                 & \textcolor{Apricot}{22.45}                & \multicolumn{1}{c|}{\textcolor{Apricot}{25.12}} & 0.767                & 0.839                & 0.882                \\

\multicolumn{1}{l|}{FreeNeRF~\cite{yang2023freenerf}}   & &        & \textcolor{Apricot}{18.02}                & \textcolor{RoyalBlue}{22.39}              & \multicolumn{1}{c|}{\textcolor{RoyalBlue}{24.20}} & \textcolor{RoyalBlue}{0.680}                & \textcolor{RoyalBlue}{0.779}                & \multicolumn{1}{c|}{\textcolor{RoyalBlue}{0.833}} & \textcolor{RoyalBlue}{19.92}                & \textcolor{RoyalBlue}{23.25}                & \multicolumn{1}{c|}{\textcolor{RoyalBlue}{25.38}} & \textcolor{RoyalBlue}{0.787}                & \textcolor{Apricot}{0.841}                & \textcolor{RoyalBlue}{0.888}                \\
\multicolumn{1}{l|}{\textbf{CVT-\textit{x}RF (w/ BARF)}}     &  &     & \textcolor{RoyalBlue}{18.06}                & \textbf{\textcolor{BrickRed}{23.40}}               & \multicolumn{1}{c|}{\textbf{\textcolor{BrickRed}{26.56}}} & \textbf{\textcolor{BrickRed}{0.762}}               & \textbf{\textcolor{BrickRed}{0.872}}                & \multicolumn{1}{c|}{\textbf{\textcolor{BrickRed}{0.910}}} & \textbf{\textcolor{BrickRed}{21.33}}               & \textbf{\textcolor{BrickRed}{25.50}}                & \multicolumn{1}{c|}{\textbf{\textcolor{BrickRed}{27.68}}} & \textbf{\textcolor{BrickRed}{0.844}}                & \textbf{\textcolor{BrickRed}{0.911}}                & \textbf{\textcolor{BrickRed}{0.938}}                \\
\bottomrule[0.5pt]

\toprule[0.5pt]
\multicolumn{1}{l|}{DietNeRF~\cite{jain2021putting}}  & \multirow{4}{*}{$\checkmark$}  &        & 10.01                & 18.70                & \multicolumn{1}{c|}{22.16} & 0.354                & 0.668                & \multicolumn{1}{c|}{0.740} & 11.85                & 20.63                & \multicolumn{1}{c|}{23.83} & 0.633                & 0.778                & 0.823                \\
\multicolumn{1}{l|}{SPARF~\cite{truong2023sparf}}     &  &\multirow{1}{*}{Optimized}    & 18.30                & 23.24                & \multicolumn{1}{c|}{25.75} & 0.780                & 0.870                & \multicolumn{1}{c|}{0.910} & 21.01                & \textbf{25.76}                & \multicolumn{1}{c|}{27.30} & 0.870                & 0.920                & 0.940                \\
\multicolumn{1}{l|}{SPARF$^{*}$}     &  &\multirow{1}{*}{per Scene}      & 18.32                & 23.43                & \multicolumn{1}{c|}{25.75} & 0.784                & 0.879                & \multicolumn{1}{c|}{0.910} & 21.26                & 25.07                & \multicolumn{1}{c|}{27.30} & 0.873                & \textbf{0.921}                & 0.940                \\
\multicolumn{1}{l|}{\textbf{CVT-\textit{x}RF (w/ SPARF)}} & & \multirow{1}{*}{} & \textbf{18.98} & \textbf{24.51} & \multicolumn{1}{c|}{\textbf{27.04}} & \textbf{0.801} &\textbf{0.884} & \multicolumn{1}{c|}{\textbf{0.919}} & \textbf{21.51} & 25.14 & \multicolumn{1}{c|}{\textbf{27.63}} & \textbf{0.874} & 0.920 & \textbf{0.945} \\ 
\bottomrule[1.pt]

\end{tabular}}
 \vspace{-8pt}
\caption{Comparison on DTU dataset. We present the performances of both full images and foreground objects. We organize the comparisons into two categories according to whether the methods use off-the-shelf models pre-trained on other datasets (indicated by \textit{Pre.}) or not. The improvement brought by the pre-trained model in RegNeRF is limited, so we place it into the first category. $^*$ means that we rerun the experiments with their official code.
The best, second-best and third-best entries of the first category of comparison are marked in \textbf{\textcolor{BrickRed}{red}}, \textcolor{RoyalBlue}{blue} and \textcolor{Apricot}{orange}, respectively. For the second category of comparison, we mark the best entries with $\textbf{bold}$. } \vspace{-5pt}\label{tab:sota_dtu}
\end{table*}

\noindent \textbf{Effect of voxel-based ray sampling.} The proposed sampling strategy is a prerequisite of the local implicit and the global explicit constraints. To study its effect, we apply our sampling strategy on NeRF~\cite{mildenhall2020nerf}, denoted as `w/o {implicit and explicit}' in Tab.~\ref{table:ablation}. It can be observed that it brings improvements over the random ray sampling strategy applied by NeRF. 
The results demonstrate that the proposed sampling strategy is more effective for the sparse-input setting. 

\par\noindent \textbf{Effect of local implicit constraint.} Tab.~\ref{table:ablation} validates that, using the local implicit constraint, denoted as `w/o {explicit}', brings a large improvement upon the voxel-based ray sampling strategy. The local implicit constraint is implemented by a Transformer architecture, which infers the radiances of the ray points from the interaction with surrounding points. 
Thus, the 3D field consistency of the region that surrounding points are distributed in is enhanced as illustrated in Fig.~\ref{fig:vis_components}. With the implicit constraint, the radiance field distribution is learned with better 3D consistency. 

\par\noindent \textbf{Effect of global explicit constraint.} As shown in Tab.~\ref{table:ablation}, the global explicit constraint also increases the performance. On DTU dataset, the PSNR is improved from 11.85 to 14.13, and the SSIM is also boosted from 0.458 to 0.518. These improvements verify that, by considering the negative pairs in the $\mathcal L_{\rm contrast}$, our method can effectively facilitate the learning of the 3D field consistency. This can also be confirmed from Fig.~\ref{fig:vis_components}: certain amounts of artifacts are removed with the global explicit constraint. 

\subsection{Performance on Different Baselines}
We validate the efficacy of our method on different baselines, which are NeRF~\cite{mildenhall2020nerf}, BARF~\cite{lin2021barf}, and SPARF~\cite{truong2023sparf}. BARF and SPARF include pose optimization modules in their methods. We switch them off for fair comparisons.

\begin{figure}[t]
	\centering
	\includegraphics[width=0.99\linewidth]{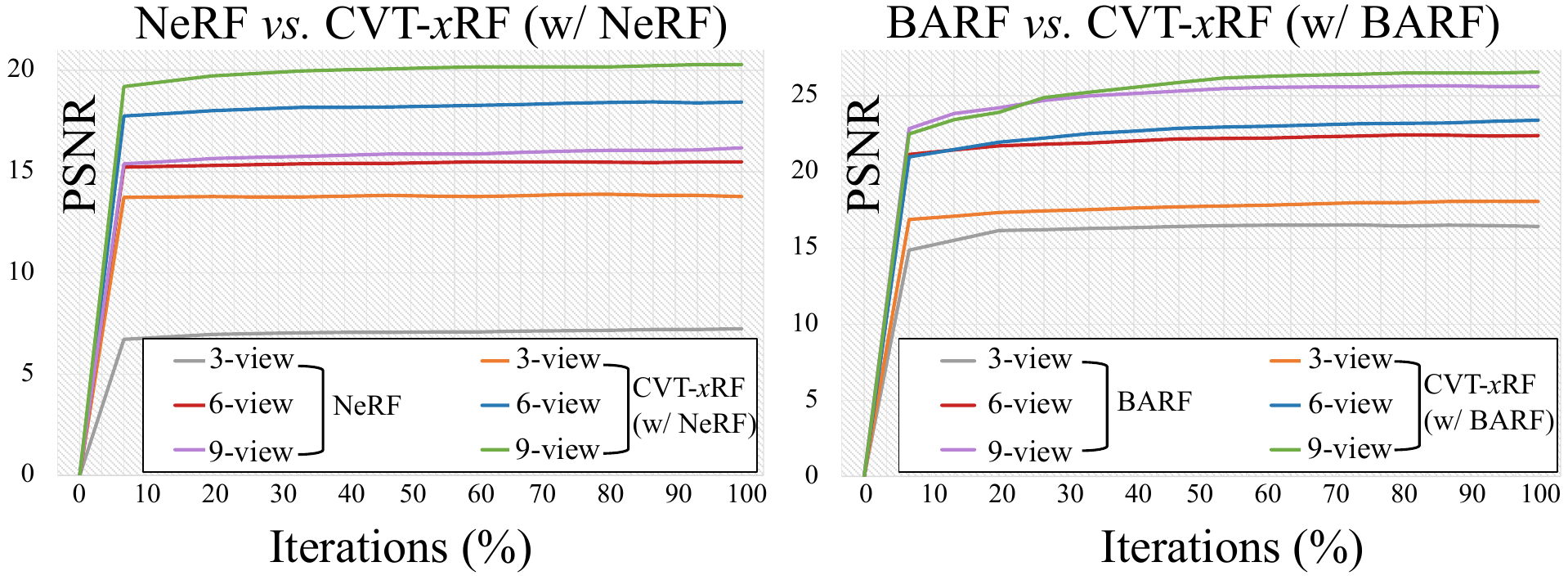}
    \vspace{-8pt}
	\caption{Performances of NeRF and BARF throughout the training process with/without the proposed CVT. } \vspace{-12pt}
	\label{fig:vis_curve}
\end{figure}

\begin{table}[h]
\centering\vspace{-3pt}
\resizebox{1.0\linewidth}{!}{
\begin{tabular}{l|c|cc|cc|cc}
\toprule[1.pt]
\multirow{2}{*}{Methods} & \multirow{1}{*}{Mem /} & \multicolumn{2}{c|}{3-view} & \multicolumn{2}{c|}{6-view} & \multicolumn{2}{c}{9-view} \\ %\cline{3-8} 
                                                    &  Time (50k)                                             & PSNR         & SSIM         & PSNR         & SSIM         & PSNR         & SSIM        \\ \midrule[0.5pt]
NeRF~\cite{mildenhall2020nerf}                                        & 4.4G / 1.6h                                           &6.68 &0.249           &15.32                         &0.626                        &16.29                        & 0.693            \\
CVT-\textit{x}RF (w/ NeRF)                            & 6.4G / 2.8h                                              & \textbf{14.13}       & \textbf{0.518} & \textbf{18.43}      & \textbf{0.713}        & \textbf{20.28}   &   \textbf{0.754}             \\ \midrule[0.5pt]
BARF~\cite{lin2021barf}                                        & 4.4G / 1.6h                                            & 16.43&  0.703&   22.26  &  0.863   &     25.54    &  0.908             \\
CVT-\textit{x}RF (w/ BARF)                       & 6.4G / 2.8h                                             & \textbf{18.06}&  \textbf{0.762} &   \textbf{23.40}   &  \textbf{0.872}  &     \textbf{26.56}   &  \textbf{0.910}            \\ \midrule[0.5pt]
SPARF~\cite{truong2023sparf}                              & $^*$20G / 11.0h                                        & 18.32 &  0.784&   23.43   &   0.879   &     25.75    &  0.910             \\
CVT-\textit{x}RF (w/ SPARF)                 & $^*$22G / 13.0h                                 & \textbf{18.98} &  \textbf{0.801}&   \textbf{24.51}   &   \textbf{0.884}   &     \textbf{27.04}    &  \textbf{0.919}             \\ \bottomrule[1.pt]
\end{tabular}
}\vspace{-8pt}\caption{Improvements and extra training overheads over different baselines on DTU dataset of different input views. $^*$ refers to peak memory consumption. The overheads are measured on $\rm{scan}40$. 
% Note that all methods above do not bring extra overheads during testing. 
}\vspace{-4pt}\label{tab:baselines}
\end{table}

\begin{figure}[th]
	\centering
	\vspace{-5pt}\includegraphics[width=0.99\linewidth]{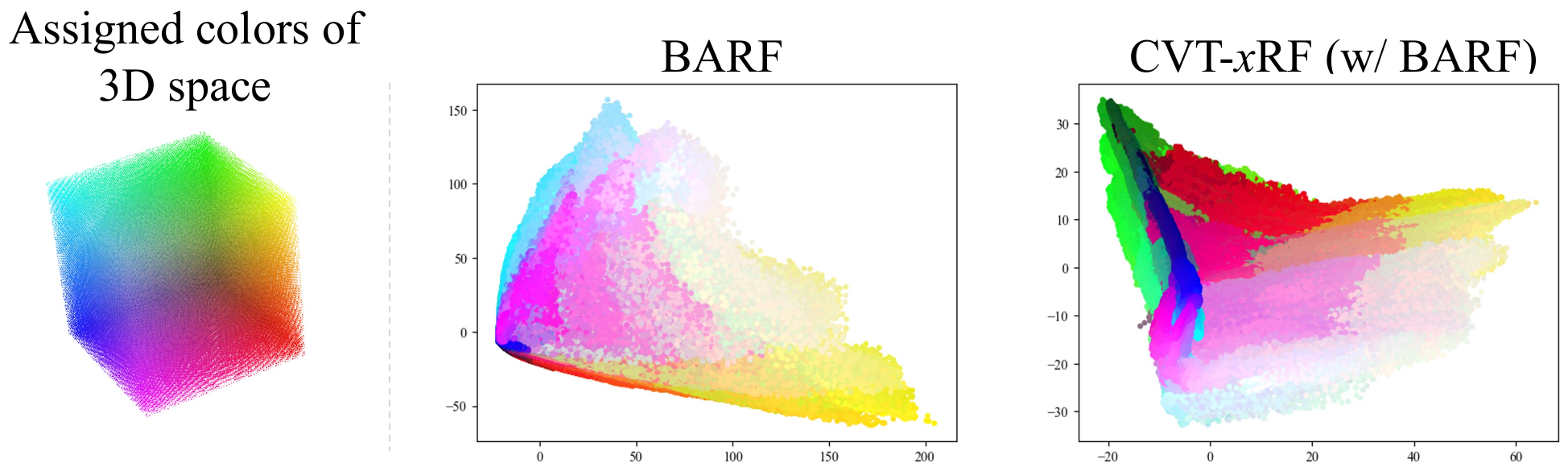}\vspace{-5pt}
	\caption{The feature distribution of DTU $\rm{scan45}$ after PCA, with and without the proposed CVT. We uniformly sample the points in 3D space and assign distinct colors to them.  It is observed that With CVT, the features demonstrate greater diversity compared to the homogeneity observed with BARF. } \vspace{-12pt}
	\label{fig:vis_pca}
\end{figure}

\noindent \textbf{Performance and training overhead.}~Tab.~\ref{tab:baselines} shows that the proposed CVT-\textit{x}RF brings considerable improvement on three baselines in all the settings. CVT-\textit{x}RF brings more than 7.0, 3.0, 3.0 PSNR improvement over NeRF given 3, 6, 9 input views, respectively. For the baselines that already achieve high performance such as BARF and SPARF, our approach can still bring a large gain regarding the different number of input views. It should be noted that SPARF is currently one of the most effective methods of learning radiance fields from sparse inputs. Tab.~\ref{tab:baselines} also shows that, our CVT-\textit{x}RF does not bring heavy overhead on different baselines, in terms of the GPU memory and the training time. 
We visualize the radiance fields learned by BARF and SPARF with and without CVT in Fig.~\ref{fig:vis_pointcloud}. It is obvious that CVT improves the 3D field consistency and effectively removes the floating artifacts in the rendered images.

\par\noindent\textbf{Convergence speed.}~The convergence speeds of models with and without the proposed CVT are illustrated in Fig.~\ref{fig:vis_curve}. In the majority of cases, the baseline models that incorporate CVT can consistently outperform the baselines throughout the entire training process. This indicates that our CVT-\textit{x}RF exhibit the capability to converge rapidly to a relatively high performance level from start of the training.

\par\noindent \textbf{Distribution of the learned MLP features.}~Fig.~\ref{fig:vis_pca} presents the distribution of features $\textbf{g}$ from the MLP learned on DTU $\rm{scan45}$. Uniform sampling of points in 3D space is conducted, and their corresponding MLP features are produced. PCA is then applied to visualize the point features. The figure demonstrates that the proposed CVT-\textit{x}RF can learn more discriminative features across different regions. This verifies the superior capability of the proposed method in modeling the scene's radiance characteristics.

\begin{figure}[t]
    \centering
	\includegraphics[width=0.99\linewidth]{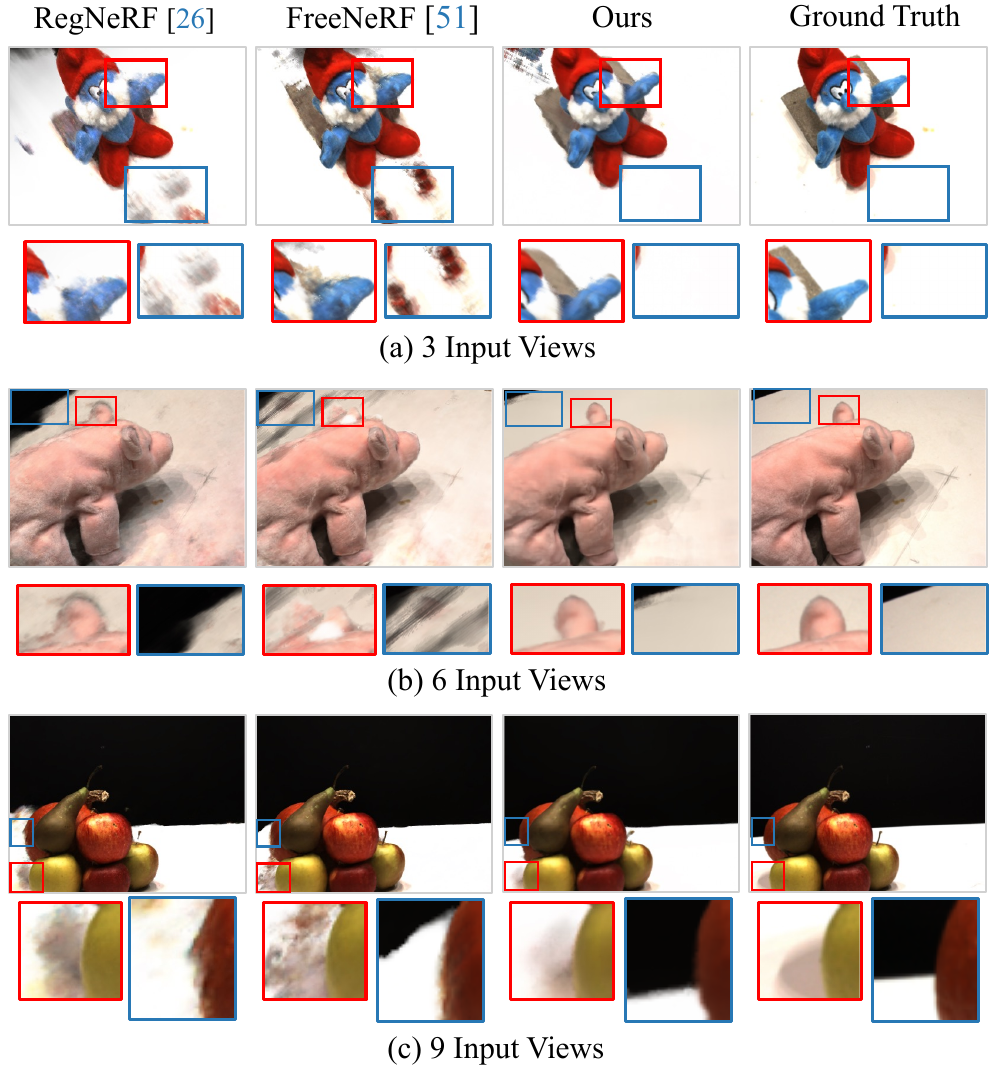}\vspace{-8pt}
	\caption{Qualitative comparisons on DTU dataset. For 3, 6, 9 input views, our method clearly preserves better consistency and exhibits significantly fewer artifacts. }
	 \vspace{-10pt}\label{fig:vis_dtu}
\end{figure}

\begin{figure}[t]
	\centering
	\includegraphics[width=0.99\linewidth]{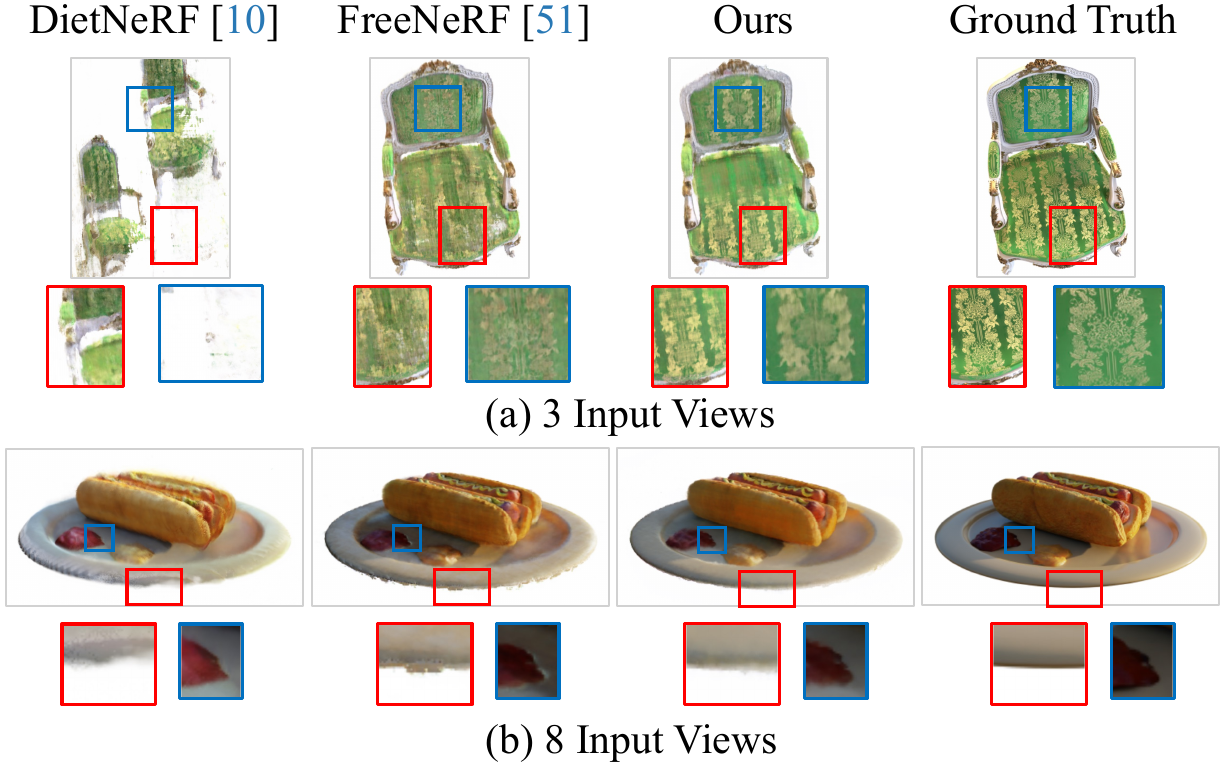}\vspace{-8pt}
	\caption{Qualitative comparisons on Synthetic dataset. For 3 input views, our method preserves more details. For 8 input views, our method shows more accurate colors and keeps sharper edges. }
	 \vspace{-12pt}\label{fig:vis_blender}
\end{figure}

\begin{table}[h]
	\begin{subtable}{0.49\linewidth} {
			\resizebox{!}{2.cm}{
				\begin{tabular}{l|cc}
					\toprule[1.pt]
					\multirow{2}{*Methods} &  \multicolumn{2}{c}{3-view}                                      \\ %\cline{2-3} 
					&  PSNR                & SSIM                               \\ \midrule[0.5pt]
					MVSNeRF~\cite{chen2021mvsnerf}                                                                     & 15.12                & 0.820                               \\
					GeoNeRF~\cite{johari2022geonerf}                                                                 & 17.67                & 0.730                                 \\
					ENeRF~\cite{lin2022efficient}                                                                    & 18.14                & 0.830                                  \\ \midrule[0.5pt]
                    %NeRF~\cite{mildenhall2020nerf}                                                                 & 11.41                & 0.724                                \\
					mip-NeRF~\cite{barron2021mip}                     & 16.52                & 0.800                                 \\ 
					DSNeRF~\cite{deng2022depth}                                                                & 15.13                & 0.820                                  \\
					DietNeRF~\cite{jain2021putting}                                                                 & 17.55           & 0.770                                   \\
                    %InfoNeRF~\cite{kim2022infonerf}                                                                 & 14.51                & 0.750                                  \\
					RegNeRF~\cite{niemeyer2022regnerf}                                                                & 17.39                & 0.820                                  \\
					%IBRNet                                                                   & 14.20                & 0.750                                  \\
					%BARF~\cite{lin2021barf}																		& 19.84         & 0.837                                \\
	                FreeNeRF~\cite{yang2023freenerf}   & \textcolor{RoyalBlue}{20.75} & \textcolor{RoyalBlue}{0.842} \\				
                    ConsistentNeRF~\cite{hu2023consistentnerf}                                               & \textcolor{Apricot}{19.63}                & \textcolor{Apricot}{0.830}                                  \\
					\small{\textbf{CVT-\textit{x}RF (w/ NeRF)}}                                                                  & \multicolumn{1}{c}{19.28} & \multicolumn{1}{c}{0.815}  \\ 
					\small{\textbf{CVT-\textit{x}RF (w/ BARF)}}                                                                  & \multicolumn{1}{c}{\textbf{\textcolor{BrickRed}{21.58}}} & \multicolumn{1}{c}{\textbf{\textcolor{BrickRed}{0.851}}}  \\ 
					
					\bottomrule[1.pt]
				\end{tabular}
			}}
        \caption{Synthetic 3-view setting.  }\label{tab:syn3view}
		\end{subtable}
		\begin{subtable}{0.49\linewidth} {
                \centering
                %\setstretch{1.7}
				\resizebox{!}{2.0cm}{
					\begin{tabular}{l|cc}
						\toprule[1.pt]
						\multirow{2}{*Methods} &  \multicolumn{2}{c}{8-view}                                      \\ %\cline{2-3} 
						&  PSNR                & SSIM                               \\ \midrule[0.5pt]
                        \multicolumn{1}{l|}{\multirow{3}{*}{\diagbox[width=10.8em,height=3.7em]{\large}{}}} & \multicolumn{2}{c}{\multirow{3}{*}{\diagbox[width=6.6
                        em,height=3.7em]{\large}{}}}  \\
                        && \\
                        && \\\midrule[0.5pt]
						NeRF~\cite{mildenhall2020nerf}                            & 14.94                & 0.687                                \\
						NV~\cite{lombardi2019neural}                             & 17.86                & 0.741                                \\
						Simplified NeRF~\cite{jain2021putting}                & 20.09               & 0.822          \\
						DietNeRF$_{50k}$~\cite{jain2021putting}                & 23.15               & \textcolor{Apricot}{0.867}          \\ 
						DietNeRF$_{200k}$~\cite{jain2021putting}               & \textcolor{Apricot}{23.59}               & \textcolor{RoyalBlue}{0.874}        \\
                        FreeNeRF~\cite{yang2023freenerf}                    & \textcolor{RoyalBlue}{24.26} & \textbf{\textcolor{BrickRed}{0.883}} \\
						\small{\textbf{CVT-\textit{x}RF (w/ BARF)}}                     & {23.33}                & \textcolor{RoyalBlue}{0.874} \\ 
                        \small{\textbf{CVT-\textit{x}RF (w/ FreeNeRF)}}                         & \textbf{\textcolor{BrickRed}{24.56}}                & \textbf{\textcolor{BrickRed}{0.883}} \\
						\bottomrule[1.pt]
				\end{tabular}}
                \caption{Synthetic 8-view setting. }\label{tab:syn8view}
				
			}
		\end{subtable}\vspace{-8pt}\caption{\small{Comparison on Synthetic dataset. The first block lists the methods that require pre-training on other datasets. The best, second-best and third-best entries are marked in \textbf{\textcolor{BrickRed}{red}}, \textcolor{RoyalBlue}{blue} and \textcolor{Apricot}{orange}, respectively. No public results for methods that require pre-training are available for 8-view. }}
    \vspace{-10pt}
    \label{tab:sota_syn}
	\end{table}

\vspace{-3pt}
\subsection{State-of-the-art Comparison}\vspace{-4pt}
In the following, unless specifically mentioned, we mainly consider BARF~\cite{lin2021barf} as our baseline and use CVT-\textit{x}RF (w/ BARF) to compare with the state-of-the-art methods. 

\par\noindent \textbf{DTU dataset. }
Tab.~\ref{tab:sota_dtu} presents the results obtained on the DTU dataset. The comparison methods are organized into two categories based on whether the methods use off-the-shelf pre-trained models on other datasets or not. 
For instance, SPARF~\cite{truong2023sparf} and DietNeRF~\cite{jain2021putting} employ a matching network~\cite{truong2021learning} and CLIP~\cite{radford2021learning}, respectively.
In the first category of comparisons, our method consistently achieves the highest performance across most cases. The qualitative results can be observed in Fig.~\ref{fig:vis_dtu}. Notably, our CVT-\textit{x}RF (w/ BARF) even achieves comparable performance to SPARF in the 6/9-view settings, while requiring significantly lower training overhead, as shown in Tab.~\ref{tab:baselines}.
Regarding the second category of comparisons, CVT-\textit{x}RF (w/ SPARF) surpasses SPARF, particularly in terms of the full-image performance. This indicates that the proposed CVT-\textit{x}RF complements the matching mechanism of SPARF that does not explicitly model the 3D field consistency.

\noindent \textbf{Synthetic dataset.}
Tab.~\ref{tab:sota_syn} shows the comparisons on the Synthetic dataset. For 3-view, our approach achieves the best results compared with other methods, and also achieves higher performance compared to the works that require pre-training~\cite{chen2021mvsnerf,johari2022geonerf,lin2022efficient}. Fig.~\ref{fig:vis_blender} (a) shows that, our method can recover better object details compared to FreeNeRF~\cite{yang2023freenerf}. For 8-view, CVT-\textit{x}RF (w/ BARF) achieves lower performance. We observe that the occlusion regularization in FreeNeRF is crucial in this setting and we combine the CVT structure with it, denoted as CVT-\textit{x}RF (w/ FreeNeRF). 
We not only achieve a 0.3 PSNR improvement over the best-performing method but also significantly enhance the radiance distribution. However, these improvements may not be evident from the evaluation metrics alone. Please refer to the supplementary material for more details.

\section{Conclusion}
In this paper, we introduce a novel approach for learning 3D spatial field consistency to regularize NeRF when training from sparse inputs. The field inconsistency is typically caused by the lack of supervision across scene views. 
We propose a novel Contrastive In-Voxel Transformer structure to learn the 3D spatial field consistency, which is composed of a voxel-based ray sampling strategy, a local implicit constraint, and a global explicit constraint. 
Our experiments demonstrate that our method outperforms various NeRF baselines in terms of 2D rendering quality, and exhibits better 3D field consistency. 
{
    \small
    \bibliographystyle{ieeenat_fullname}
    %\bibliography{main}

\begin{thebibliography}{56}
\providecommand{\natexlab}[1]{#1}
\providecommand{\url}[1]{\texttt{#1}}
\expandafter\ifx\csname urlstyle\endcsname\relax
  \providecommand{\doi}[1]{doi: #1}\else
  \providecommand{\doi}{doi: \begingroup \urlstyle{rm}\Url}\fi

\bibitem[Barron et~al.(2021)Barron, Mildenhall, Tancik, Hedman, Martin-Brualla, and Srinivasan]{barron2021mip}
Jonathan~T Barron, Ben Mildenhall, Matthew Tancik, Peter Hedman, Ricardo Martin-Brualla, and Pratul~P Srinivasan.
\newblock Mip-nerf: A multiscale representation for anti-aliasing neural radiance fields.
\newblock In \emph{ICCV}, 2021.

\bibitem[Barron et~al.(2022)Barron, Mildenhall, Verbin, Srinivasan, and Hedman]{barron2022mip}
Jonathan~T Barron, Ben Mildenhall, Dor Verbin, Pratul~P Srinivasan, and Peter Hedman.
\newblock Mip-nerf 360: Unbounded anti-aliased neural radiance fields.
\newblock In \emph{CVPR}, 2022.

\bibitem[Chen et~al.(2021)Chen, Xu, Zhao, Zhang, Xiang, Yu, and Su]{chen2021mvsnerf}
Anpei Chen, Zexiang Xu, Fuqiang Zhao, Xiaoshuai Zhang, Fanbo Xiang, Jingyi Yu, and Hao Su.
\newblock Mvsnerf: Fast generalizable radiance field reconstruction from multi-view stereo.
\newblock In \emph{ICCV}, 2021.

\bibitem[Chen et~al.(2020)Chen, Kornblith, Norouzi, and Hinton]{chen2020simple}
Ting Chen, Simon Kornblith, Mohammad Norouzi, and Geoffrey Hinton.
\newblock A simple framework for contrastive learning of visual representations.
\newblock In \emph{ICML}, 2020.

\bibitem[Deng et~al.(2022)Deng, Liu, Zhu, and Ramanan]{deng2022depth}
Kangle Deng, Andrew Liu, Jun-Yan Zhu, and Deva Ramanan.
\newblock Depth-supervised nerf: Fewer views and faster training for free.
\newblock In \emph{CVPR}, 2022.

\bibitem[Groueix et~al.(2018)Groueix, Fisher, Kim, Russell, and Aubry]{groueix2018papier}
Thibault Groueix, Matthew Fisher, Vladimir~G Kim, Bryan~C Russell, and Mathieu Aubry.
\newblock A papier-m{\^a}ch{\'e} approach to learning 3d surface generation.
\newblock In \emph{CVPR}, 2018.

\bibitem[Gu et~al.(2023)Gu, Trevithick, Lin, Susskind, Theobalt, Liu, and Ramamoorthi]{gu2023nerfdiff}
Jiatao Gu, Alex Trevithick, Kai-En Lin, Joshua~M Susskind, Christian Theobalt, Lingjie Liu, and Ravi Ramamoorthi.
\newblock Nerfdiff: Single-image view synthesis with nerf-guided distillation from 3d-aware diffusion.
\newblock In \emph{ICML}, 2023.

\bibitem[Ho et~al.(2020)Ho, Jain, and Abbeel]{ho2020denoising}
Jonathan Ho, Ajay Jain, and Pieter Abbeel.
\newblock Denoising diffusion probabilistic models.
\newblock In \emph{NeurIPS}, 2020.

\bibitem[Hu et~al.(2023)Hu, Zhou, Li, Yu, Hong, Hu, Li, Lee, and Liu]{hu2023consistentnerf}
Shoukang Hu, Kaichen Zhou, Kaiyu Li, Longhui Yu, Lanqing Hong, Tianyang Hu, Zhenguo Li, Gim~Hee Lee, and Ziwei Liu.
\newblock Consistentnerf: Enhancing neural radiance fields with 3d consistency for sparse view synthesis.
\newblock \emph{arXiv preprint arXiv:2305.11031}, 2023.

\bibitem[Jain et~al.(2021)Jain, Tancik, and Abbeel]{jain2021putting}
Ajay Jain, Matthew Tancik, and Pieter Abbeel.
\newblock Putting nerf on a diet: Semantically consistent few-shot view synthesis.
\newblock In \emph{ICCV}, 2021.

\bibitem[Jensen et~al.(2014)Jensen, Dahl, Vogiatzis, Tola, and Aan{\ae}s]{jensen2014large}
Rasmus Jensen, Anders Dahl, George Vogiatzis, Engin Tola, and Henrik Aan{\ae}s.
\newblock Large scale multi-view stereopsis evaluation.
\newblock In \emph{CVPR}, 2014.

\bibitem[Johari et~al.(2022)Johari, Lepoittevin, and Fleuret]{johari2022geonerf}
Mohammad~Mahdi Johari, Yann Lepoittevin, and Fran{\c{c}}ois Fleuret.
\newblock Geonerf: Generalizing nerf with geometry priors.
\newblock In \emph{CVPR}, 2022.

\bibitem[Kim et~al.(2022)Kim, Seo, and Han]{kim2022infonerf}
Mijeong Kim, Seonguk Seo, and Bohyung Han.
\newblock Infonerf: Ray entropy minimization for few-shot neural volume rendering.
\newblock In \emph{CVPR}, 2022.

\bibitem[Kingma and Ba(2014)]{kingma2014adam}
Diederik~P Kingma and Jimmy Ba.
\newblock Adam: A method for stochastic optimization.
\newblock \emph{arXiv preprint arXiv:1412.6980}, 2014.

\bibitem[Kosiorek et~al.(2021)Kosiorek, Strathmann, Zoran, Moreno, Schneider, Mokr{\'a}, and Rezende]{kosiorek2021nerf}
Adam~R Kosiorek, Heiko Strathmann, Daniel Zoran, Pol Moreno, Rosalia Schneider, Sona Mokr{\'a}, and Danilo~Jimenez Rezende.
\newblock Nerf-vae: A geometry aware 3d scene generative model.
\newblock In \emph{ICML}, 2021.

\bibitem[Lin et~al.(2021)Lin, Ma, Torralba, and Lucey]{lin2021barf}
Chen-Hsuan Lin, Wei-Chiu Ma, Antonio Torralba, and Simon Lucey.
\newblock Barf: Bundle-adjusting neural radiance fields.
\newblock In \emph{ICCV}, 2021.

\bibitem[Lin et~al.(2022)Lin, Peng, Xu, Yan, Shuai, Bao, and Zhou]{lin2022efficient}
Haotong Lin, Sida Peng, Zhen Xu, Yunzhi Yan, Qing Shuai, Hujun Bao, and Xiaowei Zhou.
\newblock Efficient neural radiance fields for interactive free-viewpoint video.
\newblock In \emph{SIGGRAPH}, 2022.

\bibitem[Liu et~al.(2020)Liu, Gu, Zaw~Lin, Chua, and Theobalt]{liu2020neural}
Lingjie Liu, Jiatao Gu, Kyaw Zaw~Lin, Tat-Seng Chua, and Christian Theobalt.
\newblock Neural sparse voxel fields.
\newblock In \emph{NeurIPS}, 2020.

\bibitem[Lombardi et~al.(2019)Lombardi, Simon, Saragih, Schwartz, Lehrmann, and Sheikh]{lombardi2019neural}
Stephen Lombardi, Tomas Simon, Jason Saragih, Gabriel Schwartz, Andreas Lehrmann, and Yaser Sheikh.
\newblock Neural volumes: Learning dynamic renderable volumes from images.
\newblock \emph{arXiv preprint arXiv:1906.07751}, 2019.

\bibitem[Long et~al.(2022)Long, Lin, Wang, Komura, and Wang]{long2022sparseneus}
Xiaoxiao Long, Cheng Lin, Peng Wang, Taku Komura, and Wenping Wang.
\newblock Sparseneus: Fast generalizable neural surface reconstruction from sparse views.
\newblock In \emph{ECCV}, 2022.

\bibitem[Mescheder et~al.(2019)Mescheder, Oechsle, Niemeyer, Nowozin, and Geiger]{mescheder2019occupancy}
Lars Mescheder, Michael Oechsle, Michael Niemeyer, Sebastian Nowozin, and Andreas Geiger.
\newblock Occupancy networks: Learning 3d reconstruction in function space.
\newblock In \emph{CVPR}, 2019.

\bibitem[Mildenhall et~al.(2020)Mildenhall, Srinivasan, Tancik, Barron, Ramamoorthi, and Ng]{mildenhall2020nerf}
Ben Mildenhall, Pratul~P Srinivasan, Matthew Tancik, Jonathan~T Barron, Ravi Ramamoorthi, and Ren Ng.
\newblock Nerf: Representing scenes as neural radiance fields for view synthesis.
\newblock In \emph{ECCV}, 2020.

\bibitem[M{\"u}ller et~al.(2022)M{\"u}ller, Evans, Schied, and Keller]{muller2022instant}
Thomas M{\"u}ller, Alex Evans, Christoph Schied, and Alexander Keller.
\newblock Instant neural graphics primitives with a multiresolution hash encoding.
\newblock \emph{ACM Transactions on Graphics (ToG)}, 41\penalty0 (4):\penalty0 1--15, 2022.

\bibitem[Niemeyer and Geiger(2021)]{niemeyer2021giraffe}
Michael Niemeyer and Andreas Geiger.
\newblock Giraffe: Representing scenes as compositional generative neural feature fields.
\newblock In \emph{CVPR}, 2021.

\bibitem[Niemeyer et~al.(2020)Niemeyer, Mescheder, Oechsle, and Geiger]{niemeyer2020differentiable}
Michael Niemeyer, Lars Mescheder, Michael Oechsle, and Andreas Geiger.
\newblock Differentiable volumetric rendering: Learning implicit 3d representations without 3d supervision.
\newblock In \emph{CVPR}, 2020.

\bibitem[Niemeyer et~al.(2022)Niemeyer, Barron, Mildenhall, Sajjadi, Geiger, and Radwan]{niemeyer2022regnerf}
Michael Niemeyer, Jonathan~T Barron, Ben Mildenhall, Mehdi~SM Sajjadi, Andreas Geiger, and Noha Radwan.
\newblock Regnerf: Regularizing neural radiance fields for view synthesis from sparse inputs.
\newblock In \emph{CVPR}, 2022.

\bibitem[Park et~al.(2019)Park, Florence, Straub, Newcombe, and Lovegrove]{park2019deepsdf}
Jeong~Joon Park, Peter Florence, Julian Straub, Richard Newcombe, and Steven Lovegrove.
\newblock Deepsdf: Learning continuous signed distance functions for shape representation.
\newblock In \emph{CVPR}, 2019.

\bibitem[Prinzler et~al.(2023)Prinzler, Hilliges, and Thies]{prinzler2023diner}
Malte Prinzler, Otmar Hilliges, and Justus Thies.
\newblock Diner: Depth-aware image-based neural radiance fields.
\newblock In \emph{CVPR}, 2023.

\bibitem[Qi et~al.(2017{\natexlab{a}})Qi, Su, Mo, and Guibas]{qi2017pointnet}
Charles~R Qi, Hao Su, Kaichun Mo, and Leonidas~J Guibas.
\newblock Pointnet: Deep learning on point sets for 3d classification and segmentation.
\newblock In \emph{CVPR}, 2017{\natexlab{a}}.

\bibitem[Qi et~al.(2017{\natexlab{b}})Qi, Yi, Su, and Guibas]{qi2017pointnetpp}
Charles~Ruizhongtai Qi, Li Yi, Hao Su, and Leonidas~J Guibas.
\newblock Pointnet++: Deep hierarchical feature learning on point sets in a metric space.
\newblock In \emph{NeurIPS}, 2017{\natexlab{b}}.

\bibitem[Radford et~al.(2021)Radford, Kim, Hallacy, Ramesh, Goh, Agarwal, Sastry, Askell, Mishkin, Clark, et~al.]{radford2021learning}
Alec Radford, Jong~Wook Kim, Chris Hallacy, Aditya Ramesh, Gabriel Goh, Sandhini Agarwal, Girish Sastry, Amanda Askell, Pamela Mishkin, Jack Clark, et~al.
\newblock Learning transferable visual models from natural language supervision.
\newblock In \emph{ICML}, 2021.

\bibitem[Roessle et~al.(2022)Roessle, Barron, Mildenhall, Srinivasan, and Nie{\ss}ner]{roessle2022dense}
Barbara Roessle, Jonathan~T Barron, Ben Mildenhall, Pratul~P Srinivasan, and Matthias Nie{\ss}ner.
\newblock Dense depth priors for neural radiance fields from sparse input views.
\newblock In \emph{CVPR}, 2022.

\bibitem[Rombach et~al.(2022)Rombach, Blattmann, Lorenz, Esser, and Ommer]{rombach2022high}
Robin Rombach, Andreas Blattmann, Dominik Lorenz, Patrick Esser, and Bj{\"o}rn Ommer.
\newblock High-resolution image synthesis with latent diffusion models.
\newblock In \emph{CVPR}, 2022.

\bibitem[Schwarz et~al.(2020)Schwarz, Liao, Niemeyer, and Geiger]{schwarz2020graf}
Katja Schwarz, Yiyi Liao, Michael Niemeyer, and Andreas Geiger.
\newblock Graf: Generative radiance fields for 3d-aware image synthesis.
\newblock In \emph{NeurIPS}, 2020.

\bibitem[Seo et~al.(2023)Seo, Chang, and Kwak]{seo2023flipnerf}
Seunghyeon Seo, Yeonjin Chang, and Nojun Kwak.
\newblock Flipnerf: Flipped reflection rays for few-shot novel view synthesis.
\newblock In \emph{ICCV}, 2023.

\bibitem[Sitzmann et~al.(2019{\natexlab{a}})Sitzmann, Thies, Heide, Nie{\ss}ner, Wetzstein, and Zollhofer]{sitzmann2019deepvoxels}
Vincent Sitzmann, Justus Thies, Felix Heide, Matthias Nie{\ss}ner, Gordon Wetzstein, and Michael Zollhofer.
\newblock Deepvoxels: Learning persistent 3d feature embeddings.
\newblock In \emph{CVPR}, 2019{\natexlab{a}}.

\bibitem[Sitzmann et~al.(2019{\natexlab{b}})Sitzmann, Zollh{\"o}fer, and Wetzstein]{sitzmann2019scene}
Vincent Sitzmann, Michael Zollh{\"o}fer, and Gordon Wetzstein.
\newblock Scene representation networks: Continuous 3d-structure-aware neural scene representations.
\newblock In \emph{NeurIPS}, 2019{\natexlab{b}}.

\bibitem[Sohn(2016)]{sohn2016improved}
Kihyuk Sohn.
\newblock Improved deep metric learning with multi-class n-pair loss objective.
\newblock In \emph{NeurIPS}, 2016.

\bibitem[Tatarchenko et~al.(2017)Tatarchenko, Dosovitskiy, and Brox]{tatarchenko2017octree}
Maxim Tatarchenko, Alexey Dosovitskiy, and Thomas Brox.
\newblock Octree generating networks: Efficient convolutional architectures for high-resolution 3d outputs.
\newblock In \emph{ICCV}, 2017.

\bibitem[Truong et~al.(2021)Truong, Danelljan, Van~Gool, and Timofte]{truong2021learning}
Prune Truong, Martin Danelljan, Luc Van~Gool, and Radu Timofte.
\newblock Learning accurate dense correspondences and when to trust them.
\newblock In \emph{CVPR}, 2021.

\bibitem[Truong et~al.(2023)Truong, Rakotosaona, Manhardt, and Tombari]{truong2023sparf}
Prune Truong, Marie-Julie Rakotosaona, Fabian Manhardt, and Federico Tombari.
\newblock Sparf: Neural radiance fields from sparse and noisy poses.
\newblock In \emph{CVPR}, 2023.

\bibitem[Vaswani et~al.(2017)Vaswani, Shazeer, Parmar, Uszkoreit, Jones, Gomez, Kaiser, and Polosukhin]{vaswani2017attention}
Ashish Vaswani, Noam Shazeer, Niki Parmar, Jakob Uszkoreit, Llion Jones, Aidan~N Gomez, {\L}ukasz Kaiser, and Illia Polosukhin.
\newblock Attention is all you need.
\newblock In \emph{NeurIPS}, 2017.

\bibitem[Verbin et~al.(2022)Verbin, Hedman, Mildenhall, Zickler, Barron, and Srinivasan]{verbin2022ref}
Dor Verbin, Peter Hedman, Ben Mildenhall, Todd Zickler, Jonathan~T Barron, and Pratul~P Srinivasan.
\newblock Ref-nerf: Structured view-dependent appearance for neural radiance fields.
\newblock In \emph{CVPR}, 2022.

\bibitem[Wang et~al.(2023)Wang, Chen, Loy, and Liu]{wang2023sparsenerf}
Guangcong Wang, Zhaoxi Chen, Chen~Change Loy, and Ziwei Liu.
\newblock Sparsenerf: Distilling depth ranking for few-shot novel view synthesis.
\newblock In \emph{ICCV}, 2023.

\bibitem[Wang et~al.(2021{\natexlab{a}})Wang, Liu, Liu, Theobalt, Komura, and Wang]{wang2021neus}
Peng Wang, Lingjie Liu, Yuan Liu, Christian Theobalt, Taku Komura, and Wenping Wang.
\newblock Neus: Learning neural implicit surfaces by volume rendering for multi-view reconstruction.
\newblock In \emph{NeurIPS}, 2021{\natexlab{a}}.

\bibitem[Wang et~al.(2021{\natexlab{b}})Wang, Wang, Genova, Srinivasan, Zhou, Barron, Martin-Brualla, Snavely, and Funkhouser]{wang2021ibrnet}
Qianqian Wang, Zhicheng Wang, Kyle Genova, Pratul~P Srinivasan, Howard Zhou, Jonathan~T Barron, Ricardo Martin-Brualla, Noah Snavely, and Thomas Funkhouser.
\newblock Ibrnet: Learning multi-view image-based rendering.
\newblock In \emph{CVPR}, 2021{\natexlab{b}}.

\bibitem[Wang et~al.(2004)Wang, Bovik, Sheikh, and Simoncelli]{wang2004image}
Zhou Wang, Alan~C Bovik, Hamid~R Sheikh, and Eero~P Simoncelli.
\newblock Image quality assessment: from error visibility to structural similarity.
\newblock \emph{IEEE TIP}, 13\penalty0 (4):\penalty0 600--612, 2004.

\bibitem[Warburg et~al.(2023)Warburg, Weber, Tancik, Holynski, and Kanazawa]{warburg2023nerfbusters}
Frederik Warburg, Ethan Weber, Matthew Tancik, Aleksander Holynski, and Angjoo Kanazawa.
\newblock Nerfbusters: Removing ghostly artifacts from casually captured nerfs.
\newblock \emph{arXiv preprint arXiv:2304.10532}, 2023.

\bibitem[Watson et~al.(2022)Watson, Chan, Martin-Brualla, Ho, Tagliasacchi, and Norouzi]{watson2022novel}
Daniel Watson, William Chan, Ricardo Martin-Brualla, Jonathan Ho, Andrea Tagliasacchi, and Mohammad Norouzi.
\newblock Novel view synthesis with diffusion models.
\newblock \emph{arXiv preprint arXiv:2210.04628}, 2022.

\bibitem[Wu et~al.(2015)Wu, Song, Khosla, Yu, Zhang, Tang, and Xiao]{wu20153d}
Zhirong Wu, Shuran Song, Aditya Khosla, Fisher Yu, Linguang Zhang, Xiaoou Tang, and Jianxiong Xiao.
\newblock 3d shapenets: A deep representation for volumetric shapes.
\newblock In \emph{CVPR}, 2015.

\bibitem[Yang et~al.(2023)Yang, Pavone, and Wang]{yang2023freenerf}
Jiawei Yang, Marco Pavone, and Yue Wang.
\newblock Freenerf: Improving few-shot neural rendering with free frequency regularization.
\newblock In \emph{CVPR}, 2023.

\bibitem[Yu et~al.(2021)Yu, Ye, Tancik, and Kanazawa]{yu2021pixelnerf}
Alex Yu, Vickie Ye, Matthew Tancik, and Angjoo Kanazawa.
\newblock pixelnerf: Neural radiance fields from one or few images.
\newblock In \emph{CVPR}, 2021.

\bibitem[Yu et~al.(2022)Yu, Peng, Niemeyer, Sattler, and Geiger]{yu2022monosdf}
Zehao Yu, Songyou Peng, Michael Niemeyer, Torsten Sattler, and Andreas Geiger.
\newblock Monosdf: Exploring monocular geometric cues for neural implicit surface reconstruction.
\newblock In \emph{NeurIPS}, 2022.

\bibitem[Zhang et~al.(2020)Zhang, Riegler, Snavely, and Koltun]{zhang2020nerf++}
Kai Zhang, Gernot Riegler, Noah Snavely, and Vladlen Koltun.
\newblock Nerf++: Analyzing and improving neural radiance fields.
\newblock \emph{arXiv preprint arXiv:2010.07492}, 2020.

\bibitem[Zhang et~al.(2018)Zhang, Isola, Efros, Shechtman, and Wang]{zhang2018unreasonable}
Richard Zhang, Phillip Isola, Alexei~A Efros, Eli Shechtman, and Oliver Wang.
\newblock The unreasonable effectiveness of deep features as a perceptual metric.
\newblock In \emph{CVPR}, 2018.

\bibitem[Zhou and Tulsiani(2023)]{zhou2023sparsefusion}
Zhizhuo Zhou and Shubham Tulsiani.
\newblock Sparsefusion: Distilling view-conditioned diffusion for 3d reconstruction.
\newblock In \emph{CVPR}, 2023.

\end{thebibliography}

}
\clearpage
\setcounter{page}{1}
\maketitlesupplementary

\renewcommand{\thefigure}{A\arabic{figure}}
\renewcommand{\thetable}{A\arabic{table}}
\renewcommand{\thesection}{\Alph{section}}
\setcounter{figure}{0}
\setcounter{table}{0}
\setcounter{section}{0}

% \noindent {\textit{We strongly encourage readers to refer to the accompanying demo video for a comprehensive visualization and evaluation of our method. }}

\section{Datasets and Evaluation}
\noindent \textbf{Datasets.} We evaluate our proposed method on multi-view DTU dataset~\cite{jensen2014large} and Synthetic dataset~\cite{mildenhall2020nerf}.  
DTU consists of multiple scenes. Each of them contains one or several objects on a table. We adhere to the protocol of pixelNeRF~\cite{yu2021pixelnerf} and report the performance on 15 test scenes. We report results of 3, 6, and 9 input views, following the previous PixelNeRF~\cite{yu2021pixelnerf}. Synthetic contains 8 scenes. We conduct experiments on two settings with 3 and 8 input views, respectively. For 3 input views, we follow the training and testing view selection of ConsistentNeRF~\cite{hu2023consistentnerf}. For 8 input views, we follow the view selection of DietNeRF~\cite{jain2021putting}. 

\par\noindent \textbf{Evaluation.} For quantitative comparison of synthesis results, 
beside the peak signal-to-noise ratio (PSNR) and structural similarity index (SSIM)~\cite{wang2004image} reported in the main paper, 
we also report LPIPS perceptual metric~\cite{zhang2018unreasonable} in the supplementary material. Following RegNeRF~\cite{niemeyer2022regnerf}, we additionally report the geometric mean. 
For evaluation on the DTU dataset, Niemeyer~\textit{et al.}~\cite{niemeyer2022regnerf} claim that the full image evaluation causes evaluation bias, so they mainly focus on the evaluation of object-of-interest with masks. However, we discover that a certain amount of inconsistencies in radiance fields are distributed outside objects, \textit{e.g.}, floating artifacts as illustrated in Fig.~\ref{fig:teaser}, Fig.~\ref{fig:vis_pointcloud}, and Fig.~\ref{fig:vis_dtu}. We thus report the full-image performance to test the effectiveness of our method in the main paper. For completeness, we also report the object-of-interest performance when comparing with other state-of-the-art works on the DTU dataset.

\section{Implementation Details}
\noindent \textbf{Voxel-based ray sampling. }During training, we randomly select $V$ voxels and sample $R$ rays from each of them. We set $V$ and $R$ to 64 and 16, respectively, which means we use a batchsize of 1024 rays. The entire scene is splitted into $64^3$ voxels. We set scene range of DTU dataset to 6, which means the voxel size is $(6/64)^3$. For Synthetic dataset, we set the scene range for 3 and 8 input views to 4 and 10, the voxel sizes for these two settings are thus $(4/64)^3$ and $(10/64)^3$, respectively.

\noindent \textbf{Local implicit and global explicit constraints. }The numbers of surrounding points $S$ and ray points $P$ are both set to 9. For the surrounding points sampling, we empircally set the radius of the sphere to 1/4 of the voxel size. For In-Voxel Transformer, we set the number of attention blocks to 2 for both encoder and decoder. The weight $\lambda$ of the contrastive loss is set to 0.1. 

We implement our method based on PyTorch. We train different radiance fields with the proposed CVT with Adam~\cite{kingma2014adam} optimizer. The learning rate is initially set to 5e-4 and is exponentially decayed. All experiments are performed on a single NVIDIA V100 GPU. 

\noindent \textbf{Radiance fields visualization. }For each test view, we record the density/color of each 3D point sampled on the ray during the volume rendering. The recorded points of all test views are merged into a 3D field, which is visualized by a point cloud tool. The visualized 3D fields of Fig.~\ref{fig:vis_pointcloud} differ from 2D volume-rendered synthesis, since the 3D field records the density/color of 3D points. Visualizing the 3D field helps us analyze the artifacts in the 2D images which are caused by incorrect density distribution.

\begin{table}[t]
\centering
	\begin{subtable}{0.3\linewidth} {
            \Large
			\resizebox{!}{0.9cm}{
				\begin{tabular}{c|cc}
					\hline
					$B$ &  PSNR & SSIM                               \\ \hline
					1                                                                  & 13.98                & 0.531                               \\
                    \textbf{2}                                                                  & 14.13                & 0.518                               \\
                    4                                                                  & 13.03                & 0.496                               \\
                    6                                                                  & 12.43                & 0.472                               \\

					\hline
				\end{tabular}
			}}
		\end{subtable}
  \begin{subtable}{0.3\linewidth} {
            \Large
			\resizebox{!}{0.9cm}{
				\begin{tabular}{c|cc}
					\hline
					$S$ &  PSNR & SSIM                               \\ \hline
					4                                                                  & 13.74                & 0.521                               \\
                    \textbf{9}                                                                  & 14.13                & 0.518                               \\
                    16                                                                  & 14.12                & 0.534                               \\
                    32                                                                  & 13.65                & 0.512                               \\
					
					\hline
				\end{tabular}
			}}
		\end{subtable}
    \begin{subtable}{0.3\linewidth} {
            \Large
			\resizebox{!}{0.9cm}{
				\begin{tabular}{c|cc}
					\hline
					$P$ &  PSNR & SSIM                               \\ \hline
					1                                                                  & 13.17                & 0.494                               \\
                    4                                                                  & 13.91                & 0.522                               \\
                    \textbf{9}                                                                  & 14.13                & 0.518                               \\
                    16                                                                  & 14.14                & 0.535                               \\
					
					\hline
				\end{tabular}
			}}
		\end{subtable}

		\caption{Analysis on the parameters of the Transformer. $B$, $S$ and $P$ refer to number of attention blocks, surrounding points and ray points, respectively. The parameters we select in our final model are marked in bold. }\label{tab:trans_param}
	\end{table}

\begin{figure}[t]
	\centering
	\includegraphics[width=0.99\linewidth]{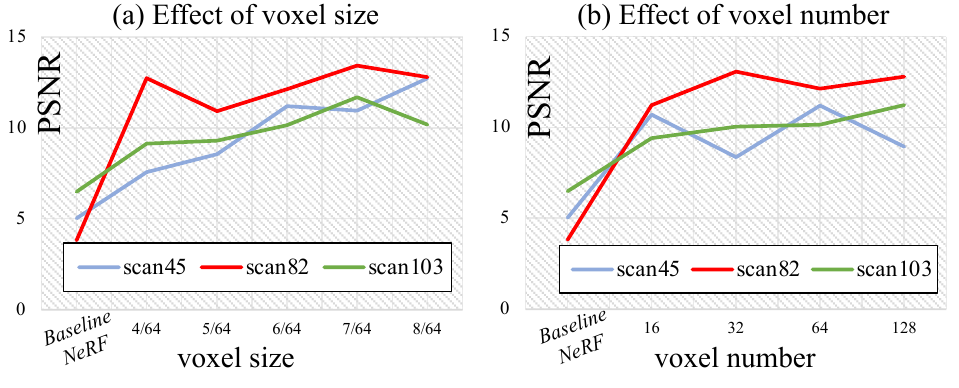}
\caption{Effects of (a) voxel size and (b) voxel number on the proposed CVT-\textit{x}RF. }
\label{fig:voxel_size}
\end{figure}

\begin{table*}[th]
\centering
\resizebox{0.98\linewidth}{!}{
\begin{tabular}{lc|c|cccccccccccc}
\toprule[1.pt]
\multicolumn{1}{l|}{\multirow{2}{*}{Methods}} & \multicolumn{1}{c|}{\multirow{2}{*}{\textit{Pre. }}} & \multicolumn{1}{c|}{\multirow{2}{*}{Setting}} & \multicolumn{3}{c|}{Full-image LPIPS$^\downarrow$}                                     & \multicolumn{3}{c|}{Full-image Average$^\downarrow$}                                     & \multicolumn{3}{c|}{Object LPIPS$^\downarrow$}                                         & \multicolumn{3}{c}{Object Average$^\downarrow$}                                    \\ 
\multicolumn{1}{l|}{}  &    & \multicolumn{1}{l|}{}            & 3-view                    & 6-view                    & \multicolumn{1}{c|}{9-view}     & 3-view                    & 6-view                    & \multicolumn{1}{c|}{9-view}     & 3-view                    & 6-view                    & \multicolumn{1}{c|}{9-view}     & 3-view                    & 6-view                    & 9-view                    \\ \midrule[0.5pt]
\multicolumn{1}{l|}{PixelNeRF~\cite{yu2021pixelnerf}}  &\multirow{4}{*}{$\checkmark$} &\multirow{1}{*}{Trained on} &0.401 &0.340 &\multicolumn{1}{c|}{0.323}  &0.154 &0.119 &\multicolumn{1}{c|}{0.105} &0.270 &0.232 &\multicolumn{1}{c|}{0.220} &0.147 &0.115 &0.100 \\
\multicolumn{1}{l|}{MVSNeRF~\cite{chen2021mvsnerf}}        & &\multirow{1}{*}{DTU and} & 0.385 &0.321 &\multicolumn{1}{c|}{0.280} &0.184 &0.146 &\multicolumn{1}{c|}{0.114}         & 0.197 &0.156 &\multicolumn{1}{c|}{0.135}& 0.113 &0.088 &0.068      \\
\multicolumn{1}{l|}{PixelNeRF ft~\cite{yu2021pixelnerf}}   &   &\multirow{1}{*}{Finetuned}  &0.456 &0.351 &\multicolumn{1}{c|}{0.338} &0.185 &0.121 &\multicolumn{1}{c|}{0.117} & 0.269 &0.223 &\multicolumn{1}{c|}{0.203} &0.125 &0.104 &0.090 \\
\multicolumn{1}{l|}{MVSNeRF ft~\cite{chen2021mvsnerf}}    &   &\multirow{1}{*}{per Scene}   &0.384 &0.319 &\multicolumn{1}{c|}{0.278} &0.185 &0.146 &\multicolumn{1}{c|}{0.113}  & 0.197 &0.155 &\multicolumn{1}{c|}{0.135} &0.113 &0.089 &\multicolumn{1}{c}{0.069}               \\ \midrule[0.5pt]
%\multicolumn{1}{l|}{NeRF~\cite{mildenhall2020nerf}}     & \multirow{5}{*}{$\times$}   & \multirow{5}{*}{Optimized per Scene}      & 6.68                     & 15.32                     & \multicolumn{1}{c|}{16.29}      & 0.249                      & 0.626                     & \multicolumn{1}{c|}{0.693}      & 7.79                 & 18.23                & \multicolumn{1}{c|}{18.80} & 0.595                & 0.758                & 0.801                \\
\multicolumn{1}{l|}{mip-NeRF~\cite{barron2021mip}}      &\multirow{4}{*}{$\times$} &     &0.655 &0.394 & \multicolumn{1}{c|}{0.209} &0.485  &0.231 & \multicolumn{1}{c|}{0.098}  &0.353 &0.198 & \multicolumn{1}{c|}{0.092} &0.323 &0.148 &0.056               \\
\multicolumn{1}{l|}{RegNeRF~\cite{niemeyer2022regnerf}}  & &\multirow{1}{*}{Optimized}        & 0.341 &0.233 &\multicolumn{1}{c|}{0.184} &0.189 &0.118 &\multicolumn{1}{c|}{0.079}  &0.190 &0.117 &\multicolumn{1}{c|}{0.089} &0.112 &0.071 &0.047              \\
\multicolumn{1}{l|}{FreeNeRF~\cite{yang2023freenerf}}   & &\multirow{1}{*}{per Scene}        & 0.318 &0.240 &\multicolumn{1}{c|}{0.187} &0.146 &0.094 &\multicolumn{1}{c|}{0.068} & 0.182& 0.137 &\multicolumn{1}{c|}{0.096} &0.098 &0.068 &\multicolumn{1}{c}{0.046}             \\
\multicolumn{1}{l|}{\textbf{CVT-\textit{x}RF (w/ BARF)}}     &  &     & \textbf{0.219}                & \textbf{0.119}              & \multicolumn{1}{c|}{\textbf{0.087}} & \textbf{0.119}            & \textbf{0.058}               & \multicolumn{1}{c|}{\textbf{0.039}} & \textbf{0.124}               & \textbf{0.072}              & \multicolumn{1}{c|}{\textbf{0.050}} & \textbf{0.071}                & \textbf{0.039}             & \textbf{0.028}               \\
\bottomrule[0.5pt]
\toprule[0.5pt]
\multicolumn{1}{l|}{DietNeRF~\cite{jain2021putting}}  & \multirow{4}{*}{$\checkmark$}  &       & 0.574 &0.336 &\multicolumn{1}{c|}{0.277} &0.383 &0.149 &\multicolumn{1}{c|}{0.098} &0.314 &0.201 &\multicolumn{1}{c|}{0.173} &0.243 &0.101 &0.068               \\
\multicolumn{1}{l|}{SPARF~\cite{truong2023sparf}}     &  & \multirow{1}{*}{Optimized}   &0.210                & 0.120                & \multicolumn{1}{c|}{0.080} & 0.113                & 0.059                & \multicolumn{1}{c|}{0.040} & \textbf{0.100}                & \textbf{0.060}                & \multicolumn{1}{c|}{\textbf{0.040}} & 0.066                & \textbf{0.036}                & \textbf{0.026}                \\
\multicolumn{1}{l|}{SPARF$^*$}     &  &\multirow{1}{*}{per Scene}      & 0.210                & 0.112                & \multicolumn{1}{c|}{0.080} & 0.113                & 0.056                & \multicolumn{1}{c|}{0.040} & 0.101                & \textbf{0.060}                & \multicolumn{1}{c|}{\textbf{0.040}} & 0.066                & 0.037                & \textbf{0.026}                \\
\multicolumn{1}{l|}{\textbf{CVT-\textit{x}RF (w/ SPARF)}} & & \multirow{1}{*}{} & \textbf{0.187} & \textbf{0.111} & \multicolumn{1}{c|}{\textbf{0.074}} & \textbf{0.102} &\textbf{0.051} & \multicolumn{1}{c|}{\textbf{0.035}} & 0.101 & 0.065 & \multicolumn{1}{c|}{0.045} & \textbf{0.063} & 0.038 & \textbf{0.026} \\ 
\bottomrule[1.pt]                   
\end{tabular}}
\caption{Comparison on DTU dataset. We present the performances of both full images and foreground objects. We organize the comparisons into two categories according to whether the methods use off-the-shelf models pre-trained from other datasets. The improvement brought by pre-trained model in RegNeRF is limited, so we place it into the first category. $^*$ means that we rerun the experiments with their official code. We mark the best entries with $\textbf{bold}$. }\label{tab:sota_dtu_lpips}
\end{table*}

\begin{table}[t]
	\begin{subtable}{0.49\linewidth} {
			\resizebox{!}{2.0cm}{
				\begin{tabular}{l|cc}
					\toprule[1.pt]
					\multirow{2}{*Methods} &  \multicolumn{2}{c}{3-view}                                      \\ 
					&  LPIPS$^\downarrow$                & Avg$^\downarrow$                               \\ \hline
					MVSNeRF~\cite{chen2021mvsnerf}                                                                     & 0.290                & 0.156                               \\
					GeoNeRF~\cite{johari2022geonerf}                                                                 & 0.330                & 0.143                                 \\
					ENeRF~\cite{lin2022efficient}                                                                    & 0.200                & 0.108                                  \\ \bottomrule[0.5pt]
					mip-NeRF~\cite{barron2021mip}                     & 0.280                & 0.141                                \\ 
					DSNeRF~\cite{deng2022depth}                                                                & 0.300                & 0.157                                  \\
					DietNeRF~\cite{jain2021putting}                                                                 & 0.280           & 0.133                                   \\
                    InfoNeRF~\cite{kim2022infonerf}                                                                 & 0.300                & 0.174                                  \\
					RegNeRF~\cite{niemeyer2022regnerf}                                                                & 0.260                & 0.126                                  \\
                    ConsistentNeRF~\cite{hu2023consistentnerf}                                               & 0.200                &  0.095                                \\
					\small{\textbf{CVT-\textit{x}RF (w/ NeRF)}}                                                                 & \multicolumn{1}{c}{0.222} & \multicolumn{1}{c}{0.104}  \\ 
                    \small{\textbf{CVT-\textit{x}RF (w/ BARF)}}                                                                  & \textbf{0.176} & \textbf{0.078}   \\ 
					\bottomrule[1.pt]
				\end{tabular}
			}
        \caption{Synthetic 3-view setting.  }\label{tab:syn3view_lpips}
        } 
		\end{subtable}
		\begin{subtable}{0.49\linewidth} 
                \centering
                %\setstretch{1.7}
				\resizebox{!}{2.0cm}{
					\begin{tabular}{l|cc}
						\toprule[1.pt]
						\multirow{2}{*Methods} &  \multicolumn{2}{c}{8-view}                                 \\
						&  LPIPS$^\downarrow$                & Avg$^\downarrow$                               \\ \midrule[0.5pt]
                        \multicolumn{1}{l|}{\multirow{3}{*}{\diagbox[width=10.8em,height=3.7em]{\large}{}}} & \multicolumn{2}{c}{\multirow{3}{*}{\diagbox[width=7.0em,height=3.7em]{\large}{}}} \\
                        && \\
                        && \\
                        \midrule[0.5pt]

						NeRF~\cite{mildenhall2020nerf}  & 0.318                & 0.122                    \\
						NV~\cite{lombardi2019neural} & 0.245                & 0.127                  \\
						Simplified NeRF~\cite{jain2021putting}  & 0.179        & 0.090          \\
						DietNeRF$_{50k}$~\cite{jain2021putting}  & 0.109         & 0.058          \\ 
						DietNeRF$_{200k}$ ~\cite{jain2021putting}  &\textbf{0.097} &  0.053     \\
                        FreeNeRF~\cite{yang2023freenerf}    &0.098    & \textbf{0.050}   \\
                        FreeNeRF$^*$ & 0.111 & 0.052 \\
						\small{\textbf{CVT-\textit{x}RF (w/ BARF)}}     & 0.116 & 0.058 \\ 
						\small{\textbf{CVT-\textit{x}RF (w/ FreeNeRF)}} & 0.108 & 0.051 \\
      
						\bottomrule[1.pt] 
                    \end{tabular}
                }
        \caption{Synthetic 8-view setting. }\label{tab:syn8view_lpips}
		\end{subtable}
		\caption{\small{Comparison on Synthetic dataset. The first block of the subtable of 3-view setting lists the methods that require pre-training on other datasets. $^*$ means that we rerun the experiments with their official code. The best entries are marked in \textbf{bold}.}}\label{tab:sota_syn_lpips}
  \vspace{-10pt}
	\end{table}

\section{Further Discussions}

\noindent \textbf{Voxel-based ray sampling. }The sampling strategy in Sec.~\ref{sec:method_sample} starts with sampling voxels in the 3D space. Note that we only sample the voxels that intersect with the training rays, removing a large number of voxels to be sampled (at least 70$\%$ removed in DTU 3-view). Though some remaining voxels correspond to the empty space, the sampled rays can can receive supervision from rendering loss, keeping training from divergence. Fig.~\ref{fig:vis_curve} shows our method achieves high performance in early training stage. 

\noindent \textbf{Local implicit constraint. }In Sec.~\ref{sec:method_implicit}, the local implicit constraint is implemented by modeling the relationship between two point sets within the voxel using a Transformer. The Transformer requires supervision from the rendering loss to ensure model convergence. Thus, the point set into the decoder has to be sampled locally along the ray, to be included in the volume rendering. This sampling is limited to line segment clipped by the voxel, i.e., the line sampling employed in Eq.~\ref{eq:ray_sample}. Sphere sampling in Eq.~\ref{eq:sur_sample} is a direct way to sample the other point set to encode local context. The sphere sampling can also be replaced by other sampling methods that can encode local context.

\section{Analysis on Parameters}
We analyze the influence of the parameters of Transformer in the proposed CVT in Tab.~\ref{tab:trans_param}. The analysis is conducted on $B$, $S$, $P$, which are number of self-attention blocks, surrounding points and ray points, respectively. We find that, as $B$ increases, the performance drops significantly, this might due to the difficulty of training the Transformer without pre-training. We also find that, the performance is slightly influenced by the number of surrounding points. We select the surrounding points to 9 as a trade-off between performance and training overhead. It is observed that, the performance increases as $P$ increases. Similarly, for the training overhead concern, we select the ray points to 9.

In the main paper, we propose a reasonable hypothesis that regions within a small voxel in 3D space are likely to present similar properties. Based on this hypothesis, we uniformly divide the scene into voxels with an equal size, and use a voxel-based ray sampling strategy in the proposed CVT-\textit{x}RF. To analyze the effect of the voxel size on the performance, we firstly fix the number of voxels that the whole scene is split into to $64\times 64\times 64$. Then we set the scene range to 4, 5, 6, 7, 8, and their respective voxel sizes are $(4/64)^3$, $(5/64)^3$, $(6/64)^3$, $(7/64)^3$, $(8/64)^3$, respectively. Fig.~\ref{fig:voxel_size} (a) shows that, though voxel size might affect the performance, CVT-\textit{x}RF brings considerable performance improvement over the baseline NeRF across a wide range of voxel sizes. 
The voxel-based sampling strategy in the proposed CVT-\textit{x}RF starts with sampling $V$ voxels for training, as stated in Sec.~\ref{sec:vbr}. We further study the effect of number of voxels. Specifically, we fix the scene range of the scene to 6 and then we experiment with the voxel number of 16, 32, 64, 128. 
As shown in Fig~\ref{fig:voxel_size} (b), CVT-\textit{x}RF brings non-negligable improvement over the baseline NeRF across a wide range of voxel number. 
From the analysis above, our method brings consistent improvement with different voxel size and voxel number, validating the effectiveness of our method.

\begin{figure*}[th]	
\centering
\includegraphics[width=0.95\linewidth]{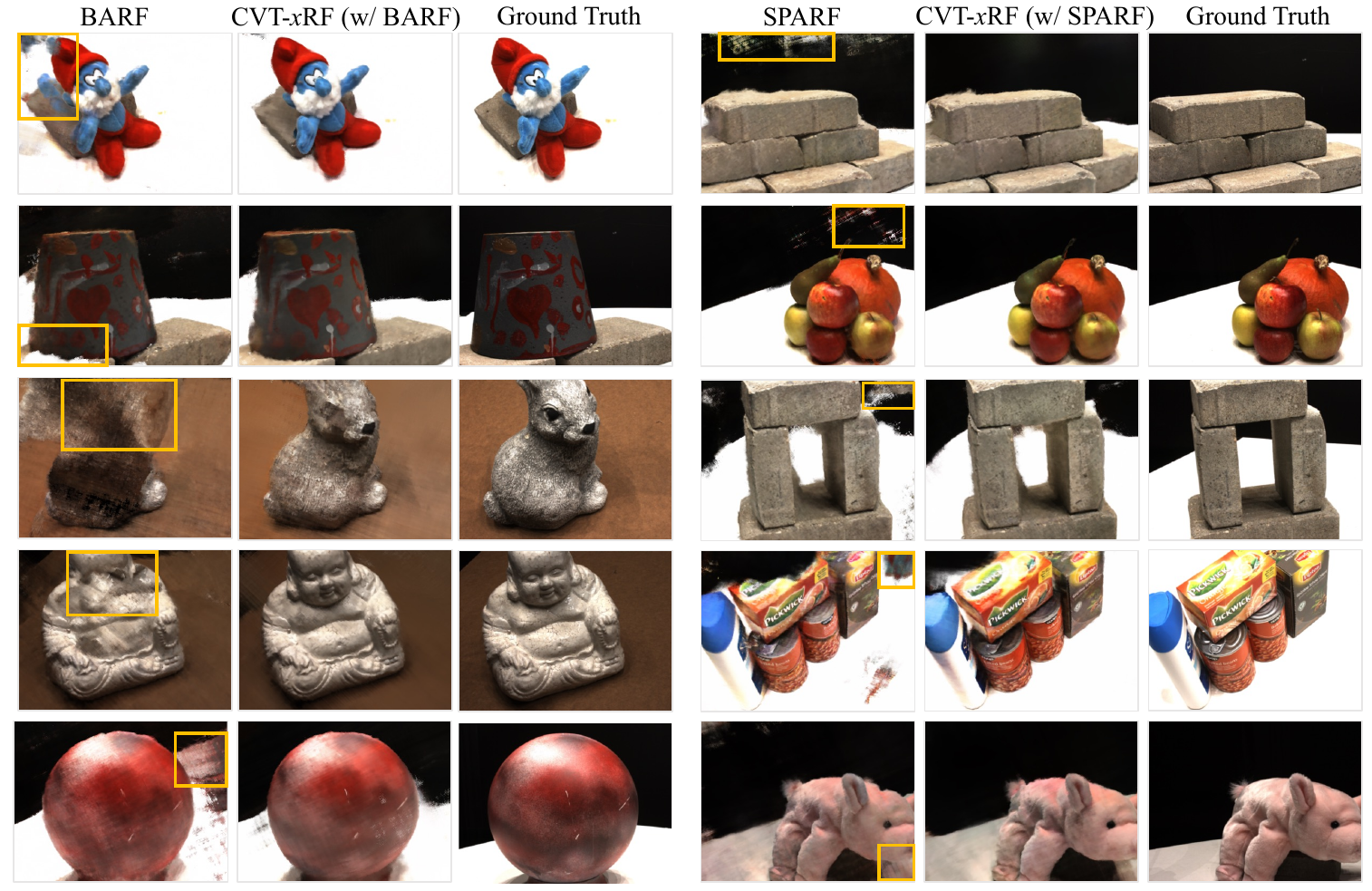}
	\caption{Qualitative comparisons between the baselines and our CVT-\textit{x}RF on the DTU dataset with 3 input views. }
\label{fig:supp_dtu_3v}
\end{figure*}

\section{Additional Quantitative Results}
Beside the quantitative comparisons in the main paper, we also compare the LPIPS perceptual metric and geometric mean~\cite{niemeyer2022regnerf} of $\rm{MSE=10^{-PSNR}}$, $\sqrt{\rm SSIM}$ and LPIPS in Tab.~\ref{tab:sota_dtu_lpips} and Tab.~\ref{tab:sota_syn_lpips}.

\noindent \textbf{DTU dataset. }As shown in Tab.~\ref{tab:sota_dtu} and Tab.~\ref{tab:sota_dtu_lpips}, our method brings significant improvement in terms of full-image evaluation. CVT-\textit{x}RF (w/ BARF) achieves state-of-the-art performance in most cases compared to those without using off-the-shelf pre-trained models. Using SPARF~\cite{truong2023sparf} as baseline, CVT-\textit{x}RF (w/ SPARF) also demonstrates improvements in PSNR/SSIM for object evaluation in most cases, although not to the same extent as observed in the full-image evaluation. This can be attributed to SPARF's reliance on the matching network for correspondence matching, which typically performs better for foreground (object) regions than for the background. 
Regarding object-level LPIPS evaluation, we observe a slight decrease when utilizing CVT for 6/9-view inputs. However, in terms of visual perception, the differences related to the object are barely noticeable, while significant visual improvements can be observed when considering the full image, as shown in Fig.~\ref{fig:supp_dtu_3v}. 
For a more comprehensive understanding, we encourage readers to refer to the demo, which provides a better visualization of the results.

\begin{figure}[th]	
    \includegraphics[width=0.9\linewidth]{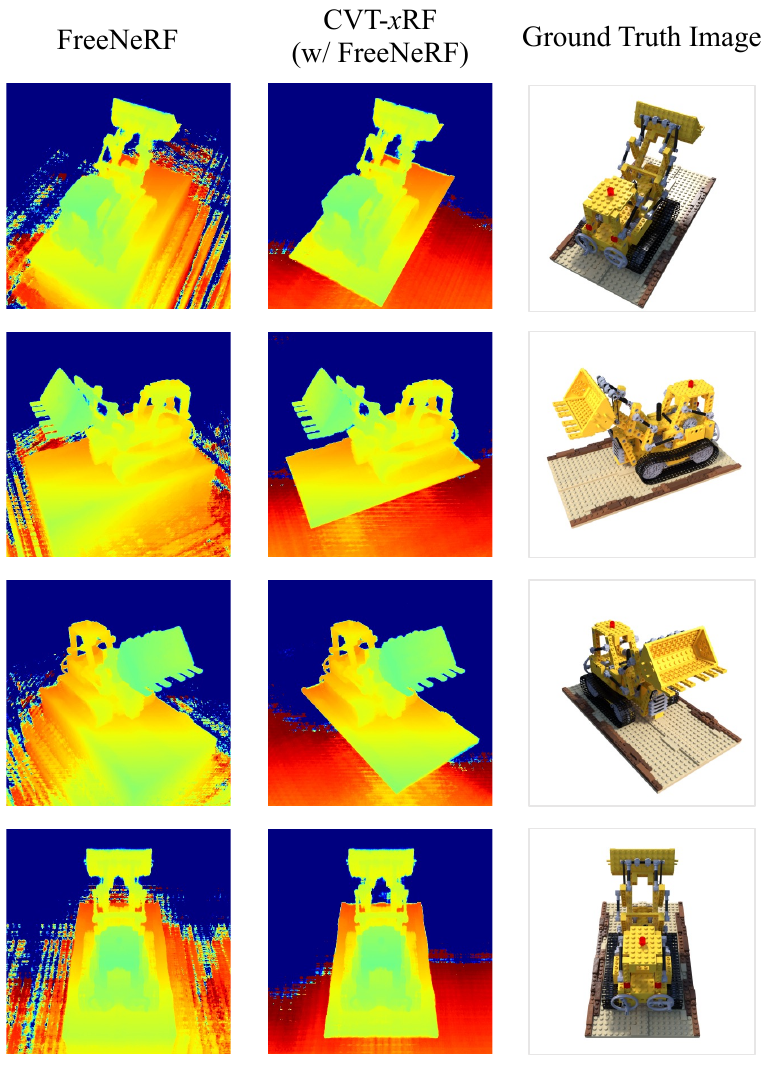}
    \centering
	\caption{Rendered depth maps on the Synthetic dataset with 8 input views. }
\label{fig:supp_blender_8v_depth}
\end{figure}

\noindent \textbf{Synthetic dataset. }
Tab.~\ref{tab:sota_syn_lpips} (a) presents the state-of-the-art performance of our method in the 3 input views setting. To evaluate our method in the 8 input views scenario, we utilize the official implementation of FreeNeRF~\cite{yang2023freenerf} (referred to as FreeNeRF$^*$) as a baseline and develop our method, named CVT-\textit{x}RF (w/ FreeNeRF). In addition to the 0.3 PSNR improvement shown in Tab.~\ref{tab:sota_syn}, our method also demonstrates an increase in LPIPS, as indicated in Tab.~\ref{tab:sota_syn_lpips} (b).
It should be noted that metrics like PSNR, SSIM, and LPIPS do not directly reflect the quality of the radiance field distribution, particularly in object-level scenes where the background is entirely blank. To gain insights into the radiance field distribution, we render depth maps for from the learned radiance field of the scene lego, as illustrated in Fig.~\ref{fig:supp_blender_8v_depth}. The results reveal a significant number of 3D points with non-negligible density values (with the color of the background) scattered throughout the free space. This leads to inaccurate depth estimation, making it challenging to determine the object geometry. In contrast, our method, which is optimized with 3D spatial field consistency, maintains clear object boundaries in terms of depth, thereby exhibiting a superior radiance field distribution.

\begin{figure*}[t]	
    \includegraphics[width=1.0\linewidth]{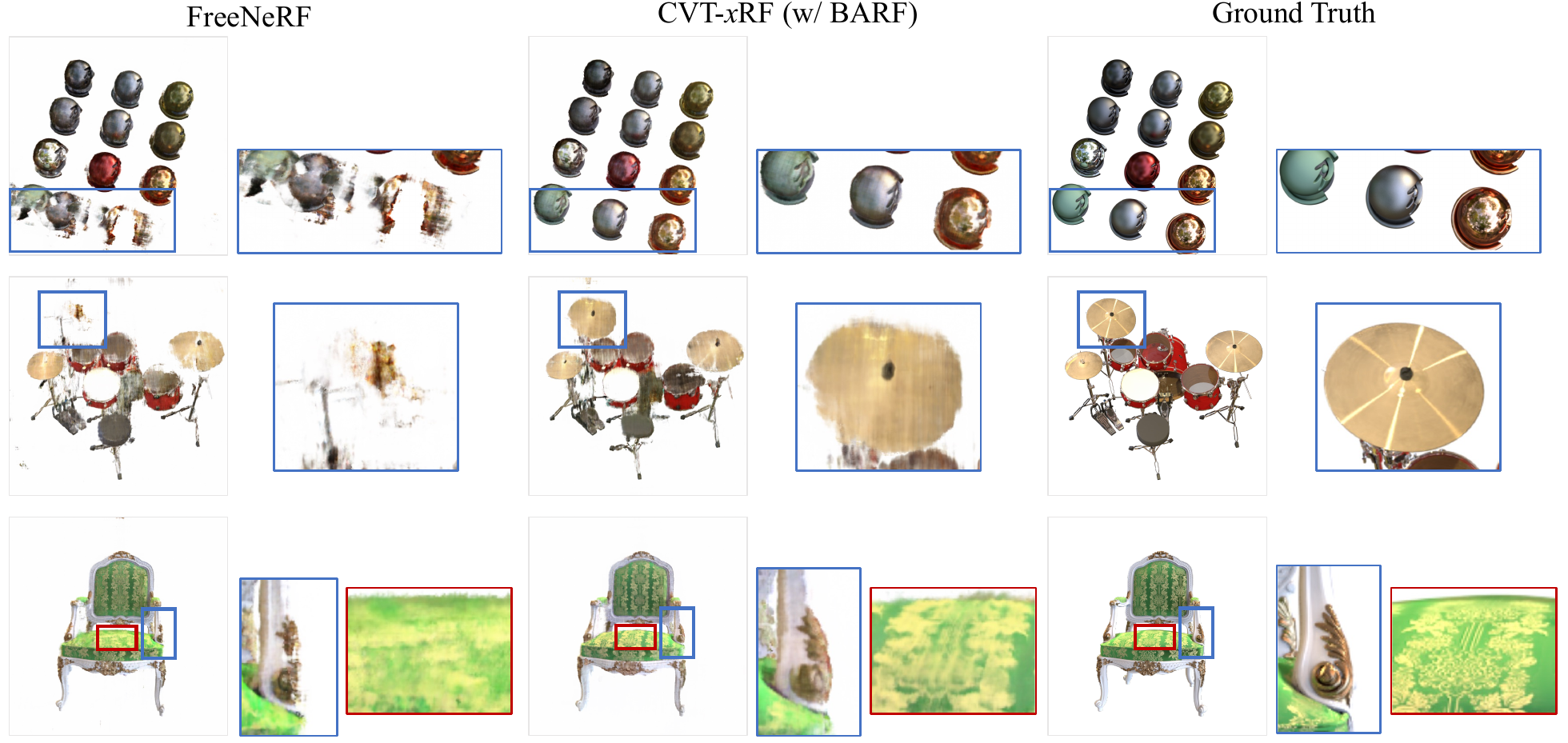}
	\caption{Qualitative comparison on the Synthetic dataset with 3 input views.  }\vspace{-1mm}
\label{fig:supp_blender_3v}
\end{figure*}

 \section{Additional Qualitative Results}

\noindent \textbf{DTU dataset. }Fig.~\ref{fig:supp_dtu_3v} visually demonstrates the qualitative improvements achieved by our method with 3 input views, using BARF~\cite{lin2021barf} and SPARF~\cite{truong2023sparf} as baselines for comparison. The comparisons reveal that our CVT-\textit{x}RF significantly reduces floating artifacts. Additionally, in certain cases where the object appears blurry, such as the third and fourth rows of BARF, CVT-\textit{x}RF (w/ BARF) better recovers the foreground objects.

 \begin{figure}[th]	
    \includegraphics[width=1.0\linewidth]{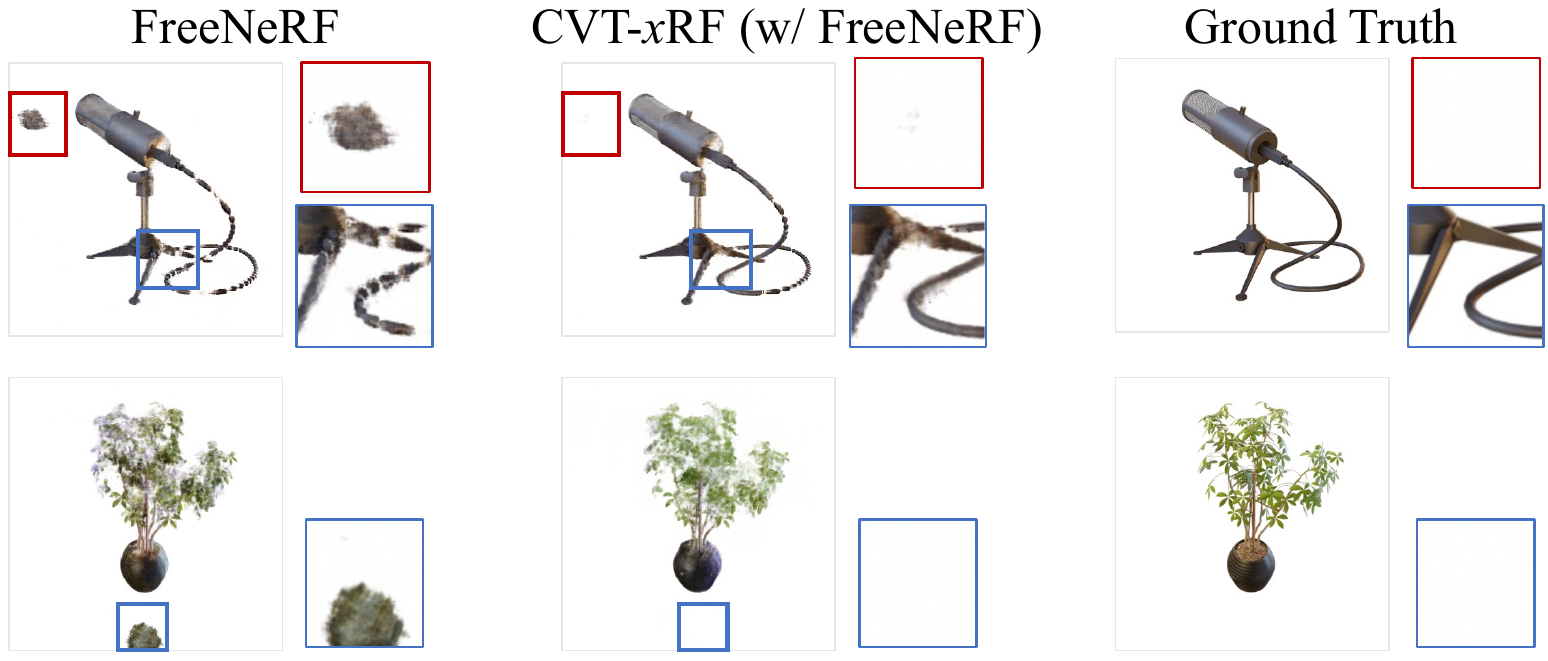}
    \centering
	\caption{Qualitative comparison on the Synthetic dataset with 8 input views. }
\label{fig:supp_blender_8v}
\end{figure}

\noindent \textbf{Synthetic dataset. }
Fig.~\ref{fig:supp_blender_3v} showcases the advantages of our method with 3 input views in terms of object completeness. Notably, our method better preserves the integrity of objects, such as the balls in the materials and the cymbal in the drums. Furthermore, our method retains finer details, as evidenced by the chair illustration.
For 8 input views, Fig.~\ref{fig:supp_blender_8v} demonstrates the superiority of our method in eliminating floating artifacts. Notable examples include the floaters in the background and the white floaters near the microphone wire.
These visual comparisons provide empirical evidence that our method's implementation of 3D spatial field consistency yields clear benefits for novel view synthesis from sparse inputs.

\end{document}